%% file: emnlp2021.tex
\pdfoutput=1
\documentclass[11pt]{article}

\usepackage[]{emnlp2021}

\usepackage{times}
\usepackage{latexsym}
\usepackage{CJKutf8}
\usepackage{multirow}
\usepackage{multicol}
\usepackage{soul}
\usepackage[hang,flushmargin]{footmisc}
\usepackage{amssymb}% http://ctan.org/pkg/amssymb
\usepackage{pifont}% http://ctan.org/pkg/pifont
\usepackage{framed}
\usepackage{color}
\definecolor{shadecolor}{rgb}{0.92,0.92,0.92}
\usepackage[textsize=scriptsize]{todonotes}
	
\usepackage{color, colortbl}

\usepackage{amsmath}
\usepackage{cleveref}
\newcommand{\Scref}[1]{\S\ref{#1}}
\usepackage{booktabs} 
\usepackage{multicol}
\usepackage{enumitem}
\usepackage{caption}
\usepackage{subcaption}
\usepackage{xcolor}
\usepackage{amssymb}
\usepackage{graphicx}
\usepackage{caption} 
\usepackage{subcaption}

\definecolor{green}{RGB}{68,169,32}
\newcommand{\greenuparrow}{\tiny\textcolor{teal}{\blacktriangle}}
\newcommand{\reddownarrow}{\tiny\textcolor{red}{\blacktriangledown}}
\usepackage[T1]{fontenc}
\usepackage[utf8]{inputenc}

\usepackage{microtype}
\title{Benchmarking Machine Translation with Cultural Awareness}
\author{Binwei Yao$^1$, Ming Jiang$^2$, Tara Bobinac$^1$, Diyi Yang$^3$, Junjie Hu$^1$ \\
  $^1$University of Wisconsin-Madison, $^2$Indiana University Indianapolis, $^3$Stanford University \\
  \texttt{binwei.yao@wisc.edu, mj200@iu.edu, 
 bobinac@wisc.edu} \\
  \texttt{diyiy@stanford.edu, junjie.hu@wisc.edu} \\
  \\ %\And
  }

\begin{document}
\maketitle
\begin{abstract}
Translating culture-related content is vital for effective cross-cultural communication. However, many culture-specific items (CSIs) often lack viable translations across languages, making it challenging to collect high-quality, diverse parallel corpora with CSI annotations. This difficulty hinders the analysis of cultural awareness of machine translation (MT) systems, including traditional neural MT and the emerging MT paradigm using large language models (LLM). 
To address this gap, we introduce a novel parallel corpus, enriched with CSI annotations in 6 language pairs for investigating \textbf{C}ulturally-\textbf{A}ware \textbf{M}achine \textbf{T}ranslation---\textbf{CAMT}.\footnote{The corpus and code are released at https://github.com/BigBinnie/Benchmarking-LLM-based-Machine-Translation-on-Cultural-Awareness} Furthermore, we design two evaluation metrics to assess CSI translations, focusing on their pragmatic translation quality. Our findings show the superior ability of LLMs over neural MTs in leveraging external cultural knowledge for translating CSIs, especially those lacking translations in the target culture. %  that LLMs have the potential to leverage cultural knowledge to enhance the understandability of translating culture-specific entities, especially those without well-known translations.
% The results of both automatic and human evaluations shed light on the cultural awareness of existing MT systems and provide valuable insights for improving the translation performance of culturally specific entities in LLM-based MT systems. \junjie{The last sentence should contain more exciting findings.}

\end{abstract}
% Recently, a new translation paradigm has emerged, which utilizes prompts to guide large language models (LLMs) to perform machine translation. This novel approach enables the seamless integration of cultural knowledge into prompt-based LLM translation. Nevertheless, a comprehensive and systematic comparison between conventional MT systems and LLM translation in terms of cultural awareness is currently lacking. 
% conduct an extensive comparison of several prompting strategies for LLM translation, aiming to comprehensively evaluate their impact on cultural awareness. 
\input{sections/01_introduction.tex}

\input{sections/07_related.tex}
\input{sections/02_data_construction.tex}

\input{sections/03_evaluation.tex}
\input{sections/04_setting.tex}
\input{sections/05_results.tex}

\input{sections/06_analysis.tex}
\input{sections/08_conclusion.tex}
\input{sections/09_limitations}
\input{sections/10_acknowledgement}

\bibliography{anthology,custom}
\bibliographystyle{acl_natbib}

\newpage
\appendix

\input{sections/00_appendix}

\end{document}

%% file: sections/01_introduction.tex
\section{Introduction}
\label{sec:introduction}
% \binwei{Outline: Cultural-aware MT is an important and under-explored area. 1. No existing cultural data; 2. hard to evaluate translation quality from the cultural side; 3. if or not previous terminology translation methods work for CSIs and LLM-based MT, what are new opportunities LLM could bring to us?}

\begin{figure}[]
    \centering   \noindent\includegraphics[width=\columnwidth]{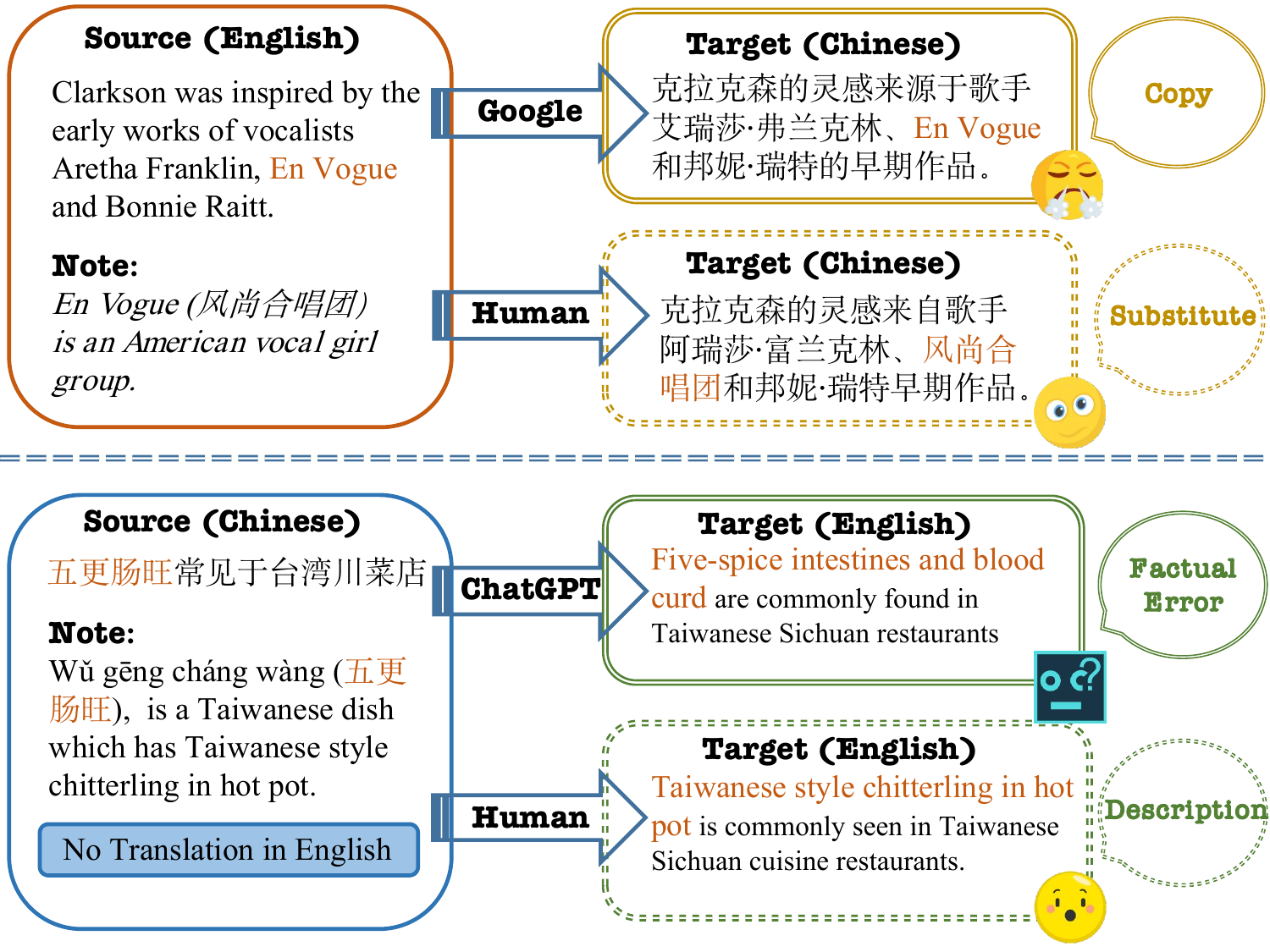}
    \caption{Culture-specific item translation errors.}
    \label{fig:examples}
    \vspace{-5mm}
\end{figure}

% \binwei{Maybe have one sentence in intro for LLM-based MT can flexibly leverage external knowledge}
Machine translation (MT) systems have achieved remarkable success in recent years, thanks in part to the pre-trained backbones of multilingual language models~\cite{aharoni-etal-2019-massively} and the availability of multilingual corpora~\cite{nllbteam2022language}. Despite these advances, terminology translation remains challenging in both general contexts~\cite{dinu2019training} and specific domains like medicine and law~\cite{ghazvininejad2023dictionary}. Many existing MT studies on terminology translation have focused on breaking language barriers rather than cultural barriers, often assuming that literal (i.e., word-for-word) translation pairs already exist for the common knowledge shared by speakers of both the source and target languages \cite{anastasopoulos2021evaluation}. However, culture is deeply intrinsic to language, and language translation entails cross-cultural communication ~\cite{newmark1988textbook, fernandez2012translating}. Due to the diverse nature of knowledge and norms across cultures, many cultural-specific items (CSIs) related to food, clothing, art, and religion are rarely used by speakers outside the items' associated cultural group, with some items even not existing in certain cultures \cite{woolford1983bilingual,persson2015culture}. As a result, cultural-specific items usually lack available literal translations across languages, leading most MT systems to perform poorly on cultural-centered translations in real-world deployment ~\cite{akinade-etal-2023-varepsilon, liebling-etal-2022-opportunities}. As shown in Figure \ref{fig:examples}, common errors, including copy and factual errors, are still made by the state-of-the-art MT systems (e.g., Google Translate and ChatGPT). More importantly, the nature of CSIs leads to difficulty in collecting high-quality, yet diverse parallel corpora at scale, hindering a systematic evaluation of both the traditional neural MT systems and the emerging MT paradigm using large language models (LLM).

To address this challenge, a handful of recent studies have begun to curate culture-related corpora for analysis from two main perspectives. The first perspective emphasizes regional varieties, such as the variety of Portuguese used in Brazil versus Portugal \cite{riley2023frmt}. The second perspective focuses on cultural content in translation, such as recipes \cite{cao2024cultural} and idioms~\cite{li2024translate}. Given the difficulty of obtaining word-for-word translation pairs for CSIs, these studies are shifting from the traditional \textit{literal translation} paradigm to \textit{free translation} which aims to convey the meaning of source texts and prioritizes readability and cultural relevance over strict accuracy and structural fidelity. Despite the valuable contributions of these works, two major concerns may still hinder the analysis of MT systems in navigating cultural nuances. First, the demand for high-quality data requires costly human annotations, restricting these studies from scaling up their curated data resources in terms of size, language pair diversity, and cultural domain coverage (see Table~\ref{tab: dataset_comparison}). Second, common MT evaluation metrics designed for literal translation lead to a lack of reliable assessments of free translation quality.

In this study, we address the two aforementioned concerns with a particular focus on translating culture-specific items. Specifically, we introduce an annotation-efficient data curation pipeline that can freely gather a diverse, large-scale, culture-centered parallel corpus while ensuring data quality. %To capture challenging examples with geo-metadata, we further conduct cultural knowledge augmentation for Cultural-Specific Items (\textbf{CSIs}). 
The resulting corpus, called \textbf{C}ulturally-\textbf{A}ware \textbf{M}achine \textbf{T}ranslation (\textbf{CAMT}), encompasses \textit{\textbf{6} language pairs, covering \textbf{6,983} CSIs across \textbf{18} concept categories from \textbf{235} countries and regions}. To facilitate automatic assessment emphasizing CSI-centered free translation, we propose two evaluation metrics: CSI-Match and pragmatic translation assessment (PTA). The CSI-Match metric evaluates the translation accuracy of isolated CSIs, independent of their sentence context. In contrast, the PTA metric assesses the comprehensibility of CSI translations within sentence contexts, emphasizing their pragmatic effectiveness in communication with native speakers of the target language. %, emphasizing the communication effectiveness of CSI translation within sentence context, we introduce the PTA metric which utilizes GPT-4 to assess the pragmatic translation quality of MT outputs according to the target culture.%, which assesses the comprehensibility of CSI translations in a reference-free manner, highlighting the MT systems' capability to adapt to the target culture.

Leveraging our CAMT corpus and designed evaluation metrics, we conduct a systematic analysis to investigate the capability of state-of-the-art neural MT and LLM-based MT systems in translating cultural content. First, we examine their efficacy with two popular terminology translation strategies that utilize the CSI dictionary. Our findings indicate that the terminology translation strategies greatly enhance the CSI translation accuracy for both neural MT and LLM-based MT systems. However, LLMs exhibit superior capability in leveraging the external dictionary compared to NMTs, particularly for CSIs that lack well-known translations. Next, to further examine LLMs' capability for integrating external knowledge into translations, we explore prompting strategies that incorporate the CSI explanations in the prompts. Our results show that incorporating CSI explanations in the prompts notably improves the pragmatic translation quality, especially for CSIs without direct translations. In summary, our contributions are as follows:
\begin{itemize}[leftmargin=13pt]\itemsep-0.2em
\item We curate a diverse parallel corpus in six language pairs with rich cultural-specific item annotations using a highly automatic pipeline. % which is constructed automatically.
\item We introduce two new evaluation metrics (CSI-Match and PTA) to assess translation quality regarding cultural nuances, particularly for terms lacking established translations.
\item We examine both LLM-based MT and NMT systems using our dataset and metrics. Our results indicate that LLMs can effectively incorporate external cultural knowledge, thereby improving the pragmatic translation quality of CSIs.
\end{itemize}

%% file: sections/07_related.tex
\section{Related Works}
\label{sec:related}

\paragraph{Culturally-Aware Machine Translation:} As languages and cultures are highly intertwined, there is a growing desire to empower cultural awareness of MT systems~\cite{hershcovich2022challenges,riley2023frmt}. However, as cultural nuances are subtle, collecting culturally sensitive data~\cite{akinade-etal-2023-varepsilon} remains costly and time-consuming. Therefore, current work on cultural-aware translations is limited to specific domains and language pairs~\cite{cao2024cultural,li2024translate}. It is also challenging to perform a human-centered evaluation of the cultural nuances~\cite{liebling-etal-2022-opportunities, li2024translate}. Existing studies have proposed strategies to evaluate cultural awareness of traditional MT systems by grounding images~\cite{khani-etal-2021-cultural}, adapting entities~\cite{peskov-etal-2021-adapting-entities} or targeting at dilates~\cite{riley2022frmt}. Different from existing culturally relevant MTs, we focus on evaluating the cultural awareness of MT by translating culture-specific items, a relatively underexplored area.  
% Previously, researchers try to incorporate terminology dictinary into translation NMT models~\cite{dinu2019training}, and LLM-based MT systems~\cite{ghazvininejad2023dictionary}, which proves to be effective for terminology translation. And there are discussions about the evaluation metric~\cite{anastasopoulos2021evaluation}. However, CSIs, which only exists in a specific cultural group, are hard to understand by other cultural groups' people. So it's not enough to evaluate them only by accuracy centered metrics, the understandablity of the words should also be taken into consideration.

% Natural language prompts has the capacity of controlling the properties of ChatGPT outputs to generated machine translation\cite{garcia2022using}. \cite{shin-etal-2022-effect}

% \cite{scao2022bloom}
\paragraph{MT with Terminology}
% Previous studies have proposed to integrate terminology dictionaries into NMT models~\cite{dinu2019training} and LLM-based MT systems~\cite{ghazvininejad2023dictionary}, proving effective for terminology translation. 
% A major thrust of existing terminology translation methods either modify the model architectures to integrate the dictionary (XXX) or feed the translation dictionary as parts of inputs to MT models~\cite{dinu2019training, ghazvininejad2023dictionary}. 
% In this study, we focus on evaluating LLM-based MT with conventional NMT methods without modifying the underlining models, as the parameters of LLMs (e.g., ChatGPT) may not be accessible.
Previous studies on machine translation with terminology focused primarily on generic domains~\cite{dinu2019training}, or popular ones (e.g., law, medicine)~\cite{ghazvininejad2023dictionary}. However, translating culture-specific items carries its own set of unique challenges because literal translations of CSIs may not exist in the target culture, making translation adaption crucial for target language readers to understand these terms \cite{vinay1995comparative}. The adaptation can create semantic asymmetry between the source words and their translations, which makes traditional translation evaluation metrics focused on semantic alignment insufficient for cross-cultural translation \cite{hershcovich2022challenges}.
% CSIs are unique to specific cultural groups, leading to difficulty in understanding the cultural nuance of CSIs for individuals from other cultures. Thus, beyond accuracy-focused metrics, assessing CSI understandability is also crucial and underexplored in the literature.%Discussions on evaluation metrics~\cite{anastasopoulos2021evaluation} have ensued. However, for CSIs unique to specific cultural groups, understanding can be challenging for individuals from other cultures. Thus, beyond accuracy-focused metrics, assessing word comprehensibility is crucial.

\paragraph{External Knowledge for MT:} There have been multiple threads of research efforts on integrating external knowledge such as bilingual translation lexicons for neural machine translation systems, including probability interpolation of lexicons~\cite{arthur-etal-2016-incorporating,khandelwal2021nearest}, data augmentation by back-translations~\cite{hu-etal-2019-domain-adaptation}, decoding with a phrase memory~\cite{wang-etal-2017-translating}, and pre-training with an entity-based denoising objective~\cite{hu-etal-2022-deep}. Despite their effectiveness, these methods require further fine-tuning of the original MTs. As the parameters of LLMs (e.g., ChatGPT) may not be accessible, we focus on tuning-free methods for integrating external knowledge in this study \cite{dinu2019training}.

\paragraph{LLM-based MT:} Large language models, such as GPT-3~\cite{gpt3}, have proven effective in machine translation for various high-resource languages~\cite{hendy2023good, jiao2023chatgpt}. In particular, a few recent studies have investigated the performance of LLM-based MT, including formality control of translation outputs~\cite{garcia2022using}, in-context translation ability during pre-training~\cite{shin-etal-2022-effect}, and multilingual translation~\cite{scao2022bloom, zhu2023multilingual}. Moreover, previous work indicates that LLMs can integrate external knowledge in the context into translation~\cite{ghazvininejad2023dictionary,li2024translate}. However, the exploration of LLM-based MT on leveraging cultural knowledge to translate culture-specific items is still lacking.
% Previously, NMT models utilized pre-training or fine-tuning techniques to incorporate external knowledge. In order to integrate cultural knowledge, such as lexicon translation, into MT systems, probability interpolation was employed ~\cite{arthur-etal-2016-incorporating,khandelwal2021nearest}. Data augmentation techniques, such as lexicon induction~\cite{hu-etal-2019-domain-adaptation}, involved conducting back-translation to extract an in-domain lexicon for pretraining out-of-domain NMT models, leading to improved performance in domain adaptation. Another approach focused on pre-training, as demonstrated by~\cite{hu-etal-2022-deep}, who designed specific pre-training tasks that utilized both monolingual data and a knowledge base to enhance the accuracy of named entity translation. However, it is important to note that both pre-training and fine-tuning methods are non-trivial and require substantial amounts of data and computing resources. 

%% file: sections/02_data_construction.tex
\section{CAMT Dataset}
\label{sec:dataset}
\begin{figure}[h]
    \center
    % \vspace{-3mm}
\includegraphics[width=0.95\columnwidth]{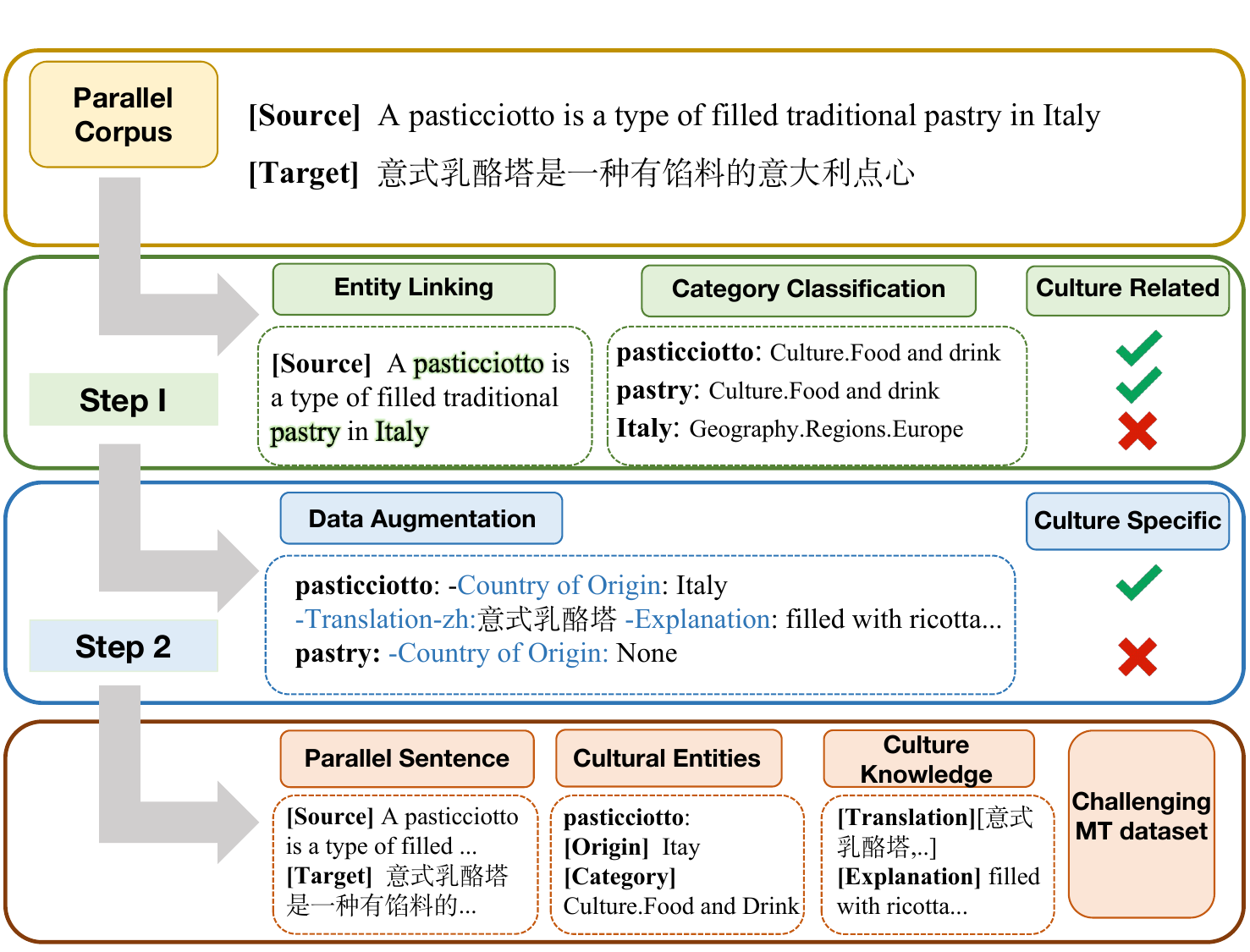}
    \caption{Overview of CAMT construction pipeline.}
    \label{fig:The Preprocessing Pipeline}
    \vspace{-3mm}
\end{figure}
% \junjie{Provide a high-level overview of the data construction pipeline. Summarize the pipeline in several steps. Finally, make a distinction between our dataset and the other existing MT dataset, and showcase our novelty.}
% In this section, we describe the construction of the culturally relevant dataset for MT. As the diverse definitions of culture and potential bias introduced by human annotators, we use Wikipedia as a knowledge base to identifying culture-specific contents. To extract culture-related Wikipedia categories, we first construct a cultural taxonomy for mapping Wikipedia categories to the classic translation theory of culture-specific items (\Scref{sec:taxonomy}). 

% A wide range of well-structured multilingual knowledge gathered by various crowd workers in Wikipedia makes it an ideal source for collecting CSIs and their descriptions in aligned languages. The overall workflow of our data collection includes three stages: (1) building a wiki-centered cultural taxonomy (\Scref{sec:taxonomy}); (2) curating parallel entities related to diverse sociocultures (\Scref{sec:parallel_text}); (3) augmenting geo-metadata (\Scref{sec:augmentation}). The curation pipeline with one example is shown in Figure \ref{fig:The Preprocessing Pipeline}. We have more data examples in Appendix \ref{sec:data_example}.

To minimize the need for human efforts while still obtaining diverse, high-quality CSI-centered translation pairs across multiple languages and cultures, we rely on Wikipedia to collect the data. The overall workflow of our data collection includes (1) building a wiki-centered cultural taxonomy (\Scref{sec:taxonomy}); (2) curating parallel sentences containing culturally-relevant entities (\Scref{sec:parallel_text}); and, (3) augmenting geo-metadata (\Scref{sec:augmentation}). Figure \ref{fig:The Preprocessing Pipeline} displays an overview of our data construction pipeline.

\subsection{Cultural Taxonomy Extraction} 
\label{sec:taxonomy}
% \junjie{describe how we extract the cultural taxonomy from the web.}
Since culture is an abstract concept, it is difficult to capture fine-grained cultural characteristics from texts directly. With this consideration in mind, we referred to an existing CSI classification framework ~\cite{newmark1988textbook}, which has been popularly used in the study of human translations of cultural concepts, to identify culturally relevant texts from Wikipedia. Specifically, there are five CSI categories in this framework, including: 1) \textit{ecology}; 2) \textit{material culture}; 3) \textit{social culture}; 4) \textit{organizations, customs, ideas}; 5)\textit{ gestures and habits}. 
% Following a primary investigation of Wikipedia texts, we find that 
Each entity-centered Wikipedia page is labeled by a variety of Wikipedia categories~\cite{asthana2018few}. To save the efforts of matching each entity on Wikipedia with each CSI category, we map CSI categories with Wikipedia categories by manually creating a mapping table (in Table \ref{tab: mapping table}) to establish connections between the two categories. Ultimately, 18 Wikipedia categories are identified as culturally related. An entity is classified as culturally related if the category of its Wikipedia page maps to one of the CSI categories.  
% The resulting mapping table and category classification tool are described in Appendix \ref{sec:map}. 
% \footnote{\url{https://en.wikipedia.org/wiki/Wikipedia:WikiProject_Categories}}
%culture-specific items~(CSI) are defined as entity words or phrases that are unique to a specific culture, and are divided into five categories: 1) \textit{ecology}; 2) \textit{material culture}; 3) \textit{social culture}; 4) \textit{organizations, customs, ideas}; 5)\textit{ gestures and habits}~\cite{newmark2003textbook}. To apply the CSI taxonomy to Wikipedia articles, we manually create a mapping table between the CSI categories and the Wikiproject categories\footnote{\url{https://en.wikipedia.org/wiki/Wikipedia:WikiProject_Categories}}~\cite{asthana2018few} by categorizing 18 culture-related Wikiproject categories into 5 CSI categories. This table allows us to use the corresponding Wikiproject categories to locate CSIs in Wikipedia texts. The mapping table is presented in Table \ref{tab: mapping table} (\ref{sec:map}).

\subsection{Culture Parallel Text Collection}
\label{sec:parallel_text}

% \begin{figure*}
%     \centering
%     \includegraphics[width=\textwidth]{assets/figures/Preprocessing2.pdf}
%     \caption{The Preprocessing Pipeline 2}
%     \label{fig:The Preprocessing Pipeline2}
% \end{figure*}

To construct a culture parallel text corpus (e.g., for English-Chinese), we collect public text articles from Wikipedia's translation tool that cover a wide range of cultural topics, and conduct sentence alignment to get parallel sentences (tools are detailed in Appendix 
\Scref{sec:parallel_corpus_collection}). 
% Specifically, we use the bilingual Wikipedia articles translated through Wikipedia's content translate tool\footnote{\url{https://en.wikipedia.org/wiki/Wikipedia:Content_translation_tool}}. This tool allows confirmed editors to translate Wikipedia articles from a source language to a target language with a machine translation system. By tracking their editing logs, we obtain the text triples consisting of the original text in a source language, the machine-translated text, and the human post-edited text in a target language. We then use a sentence alignment tool \texttt{bleu-align}\footnote{\url{https://github.com/rsennrich/Bleualign}}~\cite{sennrich-volk-2010-mt} to obtain a sentence-level parallel corpus. 
To expand the language coverage in our corpora, we also reuse open-source parallel corpora from OPUS~\cite{tiedemann2016opus}. These include Wikipedia v1.0
% \footnote{\url{https://opus.nlpl.eu/Wikipedia-v1.0.php}} 
for English to French and Spanish, as well as Samanantar v0.2
% \footnote{\url{https://opus.nlpl.eu/Samanantar-v0.2.php}} 
for English to Hindi, Tamil, and Telugu.
% Given the significant disparity in the amount of English-language corpus data available on Wikipedia compared to other languages, we additionally supplement the corpus by incorporating CWMT 2017\footnote{\url{http://nlp.nju.edu.cn/cwmt2017/guidelines.en.html}}'s Chinese to English parallel corpus, hoping to enhance the diversity of CSIs in non-English languages.
%the substantial disparity in the quantity of English as the source language corpus data on Wikipedia in comparison to other languages, in order to acquire a greater number of CSIs from languages other than English, we supplement the corpus by incorporating CWMT 2017\footnote{\url{http://nlp.nju.edu.cn/cwmt2017/guidelines.en.html}}'s Chinese to English parallel corpus CASICT2015 for English and Chinese translation pairs
To identify sentences that contain culture-related items (Step $1$ in Figure \ref{fig:The Preprocessing Pipeline}), we first perform entity-linking ~\cite{ringgaard2017sling} to identify Wikipedia entities on the source texts, then use classification tool~\cite{asthana2018few} to classify the Wikipedia categories of these entities. The categories are further mapped to our CSI categories using the cultural taxonomy (\Scref{sec:taxonomy}). The mapping table is shown in Appendix \ref{sec:map}. Finally, we only keep sentences that contain entities belonging to CSI categories. 

% \begin{itemize}
%     \item \textbf{Parallel Text Collection from Wikipedia} We collected a set of Wikipedia segments in the source language with both machine-translated and corresponding human-edited translations in the target language. Using word alignment techniques~\citep{thompson-koehn-2019-vecalign} on the collected data, we extracted triples consisting of the source text, machine-translated text, and target text.
%     \item \textbf{Entity Linking and Category Classification}
%     To filter out culturally relevant data, we performed entity-linking ~\citep{ringgaard2017sling} on the source texts, and used Wikiproject classification tool~\cite{asthana2018few} to classify the items obtained by entity-linking. We end up keeping the data that contains culturally related items.
% \end{itemize}
\subsection{Cultural Knowledge Augmentation}
\label{sec:augmentation}
Existing MT studies~\cite{arthur-etal-2016-incorporating,hu-etal-2022-deep} have used external knowledge sources (e.g., Wikidata) to improve named entity translations. To enable future adaptations of these studies on our dataset, we parse Wikidata to augment cultural knowledge of CSIs (Step $2$ in Figure \ref{fig:The Preprocessing Pipeline}), which includes their translations, descriptions, and aliases in multiple languages. We also obtain the plain text of the first paragraph of the Wikipedia article as the CSI explanation. Moreover, to identify cultural items that are specific to a certain country, we collect information on \textit{country of origin} from Wikidata for each item and remove sentences that contain only items without an associated country of origin. We refer to these groupings of data for each CSI as CSI dictionaries, an example of which is shown in the data example in Appendix \Scref{sec:data_example}. This meticulous approach enriches our dataset with supplementary cultural knowledge, enabling us to evaluate MTs' performances in handling CSIs.
\input{tables/20_comparison}
\subsection{Dataset Analysis}
% In this section, we compare our dataset with existing culturally-relevant parallel (\Scref{sec:data_comparison}) and demonstrate the characteristics of our dataset (\Scref{sec:data_characteristics}).

We briefly compare CAMT with existing datasets that similarly focus on translating culture-specific content in Table \ref{tab: dataset_comparison}. Similar to ApposCorpus and IdiomKB, translation pairs in CAMT are at the sentence level, aiming to provide fine-level textual context to explore the translation quality of CSIs from both semantic and pragmatic perspectives. Regarding data diversity, CAMT significantly expands the coverage to 18 cultural categories compared to prior work that focus on a specific domain. With respect to languages, CAMT includes 7 languages, which is more than in existing datasets ($\leq$ 4). 

\input{tables/00_data_stats}
We further conduct detailed corpus statistics on CAMT. As shown in Table~\ref{tab: dataset statistics}, our dataset contains $6,948$ parallel sentences over $6$ language pairs, of which $3,029$ sentences have CSI translations and the rest are non-translation instances. The total number of unique CSIs (called CSIs Types) in CAMT is $6,983$. Among various cultural categories, we find that \textit{organizations, customs, ideas} and \textit{material culture} are the top 2 categories, and \textit{social culture} and \textit{ecology} are the bottom ones. A more detailed breakdown of statistics of CAMT can found in Appendix \Scref{sec:stats}.

%% file: tables/20_comparison.tex
\begin{table}[t]
    \centering
    \resizebox{\linewidth}{!}{
    \begin{tabular}{c|lcc}
    \toprule
    \textbf{Dataset}  &\textbf{Language}  & \textbf{Domains} & \textbf{Format} \\
    \midrule
    ApposCorpus  & en, es & Person  & Sent. \\ \cite{kementchedjhieva2020apposcorpus} & de, pl & Organization\\
     Adaption  & en, de & Celebrity & Ent. \\
    \cite{peskov2021adapting}\\
    IdiomKB & en, zh & Idiom & Sent. \\
    \cite{li2024translate} & ja &&\\
    CulturalRecipes  & en, zh & Recipes & Para. \\
    \cite{cao2024cultural}\\
     \midrule
     \multirow{3}{*}{\textbf{CAMT (Ours)}} & en, zh,  & Cultural  & Sent. \\
     & fr, es & Categories & \\
     & hi, ta, te & \\
    \bottomrule
    \end{tabular}
    }
    \vspace{-3mm}
    \caption{Dataset comparison. Sent., Ent., and Para. are abbreviations of sentence, entity and paragraph.}
    \label{tab: dataset_comparison}
    \vspace{-5mm}
\end{table}

%% file: tables/00_data_stats.tex
% \begin{table}[h]
%     \centering
%      \resizebox{0.6\linewidth}{!}{
%     \begin{tabular}{ll}
%     \toprule
%        \textbf{Data} & \textbf{Count} \\
%      \midrule 
%      Parallel sentences  & 1,140  \\
%      Countries (Region)  & 21  \\
%      Wikiproject Categories & 18  \\
%      CSIs~(Total) & 1,148 \\
%      CSIs~(Different)  & 537 \\
%      CSIs~(with translation)& 896 \\
%      \bottomrule
%     \end{tabular}}
%     \caption{Dataset Statistics}
%     \vspace{-4.5mm}
%     \label{tab:dataset statistics}
% \end{table}
\begin{table}[b]
    \centering
    \vspace{-3mm}
     \resizebox{\linewidth}{!}{
    \begin{tabular}{lrrrr}
    \toprule
       \textbf{Pair} & \textbf{Sent.} & \textbf{CSIs~Counts} & \textbf{CSIs Types} & \textbf{CSI Translations} \\
       \midrule
       En-Zh & 778 & 794 & 601 & 730 \\
       En-Fr &  2,073 & 2,213 & 2,213 & 1,130 \\
       En-Es & 1,580 & 1,652 & 1,652 & 817  \\
       En-Hi & 1,086 & 1,127 & 1,127 & 168 \\
       En-Ta & 677 & 695 & 695 & 118 \\
       En-Te & 754 & 695 & 695 & 66 \\
       Total & 6,948 & 7,176 & 6,983 & 3,029\\
     % \midrule 
     \bottomrule
    \end{tabular}}
    \vspace{-3mm}
    \caption{Dataset Statistics on Six Language Pairs.}%\junjie{what do you mean by CSI (different) and CSI (translated)?}}
    % \vspace{-3mm}
    \label{tab: dataset statistics}
\end{table}

%% file: sections/03_evaluation.tex
\section{Cultural Awareness Evaluation}
\label{sec:culture_mt_evaluation}
% \binwei{CSI-Match: evaluate accuracy in the context, with several candidates; Understandability: evaluate understandability beyond semantic similarity}
% Existing evaluation methods for terminology machine translation primarily focused on assessing the adequacy and fluency of the translated terms~\cite{anastasopoulos2021evaluation}. However, CSI usually has more than one translation using various strategies, leading to challenges of fine-grained evaluation.
To better capture the cultural nuances in CSI translations, we devise two evaluation metrics: (1) CSI-Match, which evaluates the accuracy of CSIs with labels, and (2) PTA, which assesses the pragmatic translation quality of CSIs without labels.

\paragraph{CSI-Match:} Existing evaluation metrics for terminology are efficient for evaluating the accuracy of translations and the fluency of outputs \cite{anastasopoulos2021evaluation}. However, previous metrics such as Exact Match (EM) assume that terminology translations must be exact matches, while reasonable adaptations of CSIs are also acceptable \cite{vinay1995comparative}. To address this, we introduce the CSI-Match metric as a modification to the EM evaluation metric. CSI-Match measures the accuracy of term translation using a more nuanced, fuzzy matching approach. It calculates the maximal partial similarity ratio (PSR) between the reference CSI translations ${t_1, t_2, \ldots, t_n}$ and the system output sentence $S$. CSI-Match is determined by Eq. \eqref{eq: CSI-Match}, resulting in a value from 0 to 100. A higher value indicates a stronger similarity of the predicted CSIs to a set of CSI translation references.
\begin{equation}
    \begin{aligned}
     \text{CSI-Match} = \max_{t \in \{t_1, t_2, \ldots, t_n\}} \text{PSR} (t, S)
    \end{aligned}
    \label{eq: CSI-Match}
\end{equation}
PSR measures the maximum similarity between string $t$ and any substring in $S$. 
% To compute PSR, we utilize the string matching tool \texttt{FuzzyWuzzy}\footnote{\url{https://github.com/seatgeek/fuzzywuzzy}}.
\begin{align} \label{eq: psr}
&\text{PSR}(t, S) =  \max_{s \in P}  (1-d(t, s))\times 100 \\  
%\in & [0,1])\\
 &P = \{ S_{i:j} \mid 0 \leq i \leq j < |S| \}
\end{align}
where $S_{i:j}$ is a substring of $S$ from word index $i$ to $j$, and $d(,)$ is the normalized Levenshtein distance which measures the minimum number of insertions, deletions, and substitutions required to change one string into another ~\cite{levenshtein1966binary}.

\paragraph{Pragmatic Translation Assessment (PTA):} Existing evaluation metrics like BLEU \cite{papineni2002bleu} and COMET \cite{rei2020comet} mainly focus on surface-level or semantic-level accuracy between the source and target texts. However, in the context of CSI translation where translation quality is tightly associated with target culture \cite{hershcovich2022challenges}, pragmatic translation quality becomes crucial. For example, a free translation based on the description of the CSI might have better pragmatic translation quality than a direct literal translation, as it would be easier for people from the target culture to understand. Therefore, we design a new assessment metric called \textit{PTA}, measuring the win rate at which CSI translations by the MT system are judged to exhibit better pragmatic translation quality compared to human reference translations. Specifically, in the human evaluation, we specify what the CSIs are in the source language and ask native speakers of the target language to compare the CSI translations within the sentential context between an MT system and a reference, and then select the translation that is easier to understand. To improve the applicability of PTA when native speakers are not available, we use GPT-4o to replace human judgments for PTA, which has proven effective for the automatic evaluation of generative models in recent studies~\cite{rafailov2023direct, kocmi2023large}. The evaluation prompt, shown in Appendix \Scref{sec:evaluation_prompts}, is used for both human annotators and GPT-4. A higher PTA score means the system translates the CSI in a more comprehensive manner than the reference sentence does, which might use other accurate but less understandable translations of CSIs.

%% file: sections/04_setting.tex
\input{tables/16_prompting_examples}
\section{Experimental Settings}
\label{sec: setting}
To fully evaluate the efficacy of MT system translating CSIs, we compare NMT systems with LLM-based MTs (\Scref{sec: mt_systems}). Secondly, to investigate the performances of traditional terminology translation methods on CSI translations, we evaluate two dictionary-based terminology methods on open-sourced NMT and LLMs (\Scref{sec:dict_method}). Moreover, to gauge the capacity of LLMs to leverage external knowledge, we evaluate 4 cultural knowledge prompting strategies on tuning-free LLMs (\Scref{sec:culture_mt}).

\subsection{MT systems in Comparison}
\label{sec: mt_systems}
We evaluate the following MT systems: 
\begin{itemize}[leftmargin=13pt]\itemsep-0.2em
    \item \textbf{NMTs}: We asses the NLLB 1.3B~\cite{nllbteam2022language} model, which is a state-of-the-art multilingual MT model.  We also use the Google Translate engine in our comparison.
    \item \textbf{LLMs}: We examine ChatGPT (GPT-3.5-turbo-1106) and the open-source LLaMA2-7B for comparison, as both have been proven to be efficient multilingual MT tools \cite{zhu2023multilingual}.
\end{itemize}

\subsection{Dictionary-based Methods in Comparison}
\label{sec:dict_method}
For the open-sourced models (i.e., LLaMA and NLLB) we experiment with two additional methods proven to be highly effective in terminology MT during inference~\cite{dinu2019training}. Specifically, we employ 1) the \textbf{Append} method: append the CSI dictionary before the input, whose format is \ul{``<CSI$_{1}$>:<CSI$_{1}$ Translations>,...,<CSI$_{n}$>:<CSI$_{n}$ Translations>[Source language]''}; and 2) the \textbf{Replace} method: replace the CSIs in the source sentence with their translation in target language. For LLaMA2, we use the following prompt in 8 shots: \ul{[Source language]:[Source sentence] = [Target language]:[Target sentence]}, which is efficient for machine translation \cite{zhu2023multilingual}.

% Specifically, we examine the vanilla \textbf{NLLB} model and its two variations by Append (\textbf{NLLB-A}) and Replace (\textbf{NLLB-R}). For open-sourced LLMs, we use \textbf{LLaMA2}~(7B) with in-context learning to perform machine translations. For each language pair, we provide 8-shot in-context exemplars using the format \ul{``[Source language]:[Source sentence]=[Target language]:[Target sentence].''} The two variants targeting CSI translations are: 1) \textbf{LLaMA2-A} that \textit{appends} the CSI translation lexicon to the source sentence in the prompt, whose format is \ul{``<CSI$_{1}$>:<CSI$_{1}$ Translations>,...,<CSI$_{n}$>:<CSI$_{n}$ Translations>[Source language]:[Source sentence] = [Target language]:[Target sentence]''}; and 2) \textbf{LLaMA2-R} that \textit{replaces} the CSIs in the source sentence with their target translation. 

\subsection{Prompting Strategies in Comparison}
\label{sec:culture_mt}
We explore various prompting strategies to introduce cultural knowledge into LLM-based MT. 
% A prompt for LLMs typically consists of a natural language task description accompanied by few-shot in-context examples. 
Our strategies generate in-context examples to integrate additional cultural knowledge, which involves employing CSI dictionaries and CSI explanations. Table~\ref{tab: Prompting Strategy} shows examples of four prompting strategies.
% Additionally, we delve into the prompting strategies eliciting from LLMs' \textit{internal knowledge}. 

%\junjie{This will be the main technical section to show the prompting methods that we've tried in this study. }

\begin{itemize}[leftmargin=13pt]\itemsep-0.2em
\item \textbf{Basic Instruction (BI)} The basic machine translation prompt of LLMs (e.g. ChatGPT).
\item \textbf{External CSI Translation (CT)}
\label{sec:ent_translate}
To assess the impact of incorporating a CSI dictionary within the prompts, we include CSIs along with their corresponding translations prior to the basic translation instruction BI. 
\item \textbf{External CSI Explanation (CE)}
\label{sec:ent_explain}
CSIs may not have a direct equivalence in the target language's culture. Therefore, it becomes necessary to translate based on the explanation of the CSI to assist the target audience in better understanding the content. To assess the impact of explanations, we include the CSI explanation obtained from Wikipedia in the prompt before the basic translation instructions.
\item \textbf{Self-Explanation (SE)}
We also examine LLMs' internal knowledge for explaining the meaning of CSIs. Inspired by Chain-of-Thought (CoT)~\cite{wei2022chain,kojima2022large}, we treat the explanation of CSIs in a source sentence as an intermediate reasoning step before translating the whole sentence. We design the explanation prompting strategy in two steps for machine translation. First, we prompt the LLM to explain the meaning of all CSIs in the source sentence.
Second, we ask the LLM to translate the whole sentence by combining the LLM's explanation with another prompt instruction. 
\end{itemize}

%% file: tables/16_prompting_examples.tex
\begin{table*}[h]
    \centering
    \resizebox{\linewidth}{!}{
    \begin{tabular}{ll}
    \toprule
    % \textbf{Source} & \multicolumn{2}{l}{alongside other Italian pastries \& cannoli like \textbf{cannoli} and \textbf{sfogliatelle} }\\
    % \textbf{Reference} & \multicolumn{2}{l}{alongside other Italian pastries \junjie{Replace this to human translation}}\\ 
    % \midrule \midrule
    \textbf{Strategy}     &  \textbf{Prompt} \\
    \midrule
\multirow{1}{*}{Basic Instruction (\textbf{BI})} & {Translate the following English text to Chinese}\\
\midrule

\multirow{3}{*}{CSI Translation (\textbf{CT})} & The Chinese translation of culture entities in the sentence is as following:\\
& {\color{blue}{\textit{cannoli:\begin{CJK}{UTF8}{gbsn}里考塔芝士卷(\textit{Ricotta cheese rolls}), 奶油甜馅煎饼卷 (\textit{Sweet Cream pancake rolls})\end{CJK}}}}\\
&Translate the following English text to Chinese \\
\midrule

\multirow{3}{*}{CSI Explanation (\textbf{CE})} & {The explanation of culture entities in the sentence is as following:} \\
&{\color{blue}{\textit{Cannoli are Italian pastries consisting of tube-shaped shells of fried pastry dough ...}}} \\
&Translate the following English text to Chinese\\
\midrule

\multirow{3}{*}{Self-Explanation (\textbf{SE})} & User: Please explain cannoli in {\color{blue}{[Source Sentence]}} \\
&LLM: {\color{blue}{[Explanation]}}\\
&User: According to your explanations to cannoli, only translate the following English text to Chinese\\
% \midrule

% \multirow{5}{*}{Self-Correction (\textbf{SC})} & {User: Translate the following English text to Chinese: {\color{blue}{[Source Sentence]}}.}& \begin{CJK}{UTF8}{gbsn}和其他意大利糕点，如卡诺利\end{CJK}\\
% &  {\color{blue}{\textit{Please firstly think about how to ensure that the culturally relevant words in the}}} &  \begin{CJK}{UTF8}{gbsn}和卷层蛋糕一起，享用\end{CJK}\\
% & {\color{blue}{\textit{sentence are translated into Chinese words that Chinese readers can understand.}}} & \\
% & LLM: \color{blue}{[Analysis]} & \\
% & User: According to your analysis above, please translate the sentence& \\
% \midrule

% \multirow{5}{*}{Self-Ranking (\textbf{SR})} & {User: Translate the following English text to Chinese: {\color{blue}{[Source Sentence]}}.}\\
% &  {\color{blue}{\textit{Please give [Generated Number] most likely translations, and ensure ''cannoli'' in each result to correspond to different translations }}} \\
% & LLM: \color{blue}{[Translations]}  \\
% & User: \color{blue}{\textit{Please select the best translation result of entities: cannoli. The translation of entitis in the sentence}} \\
% &\color{blue}{\textit{should be the closest to its explanation, and is easy to be understood by Chinese readers.}} \\
 \midrule \midrule   
    \textbf{Source} & \multicolumn{1}{l}{They are also commonly available at Italian-American bakeries in the United States,} \\
    & alongside other Italian pastries like {\color{red}{cannoli}} and {\color{red}{sfogliatelle}}.\\
     \textbf{Knowledge} & Translations:{\color{blue}{cannoli:\begin{CJK}{UTF8}{gbsn}里考塔芝士卷(\textit{Ricotta cheese rolls}), 奶油甜馅煎饼卷 (\textit{Sweet Cream pancake rolls})\end{CJK}}}\\
    & Explanation: \color{blue}Cannoli are Italian pastries consisting of tube-shaped shells of fried pastry dough ... \\
    % \textbf{Reference} & \multicolumn{1}{l}{\begin{CJK}{UTF8}{gbsn}在美国,它们通常也可以在意大利裔美国人开的面包店中见到，就像其它意大利糕点\end{CJK}}\\
    % &\begin{CJK}{UTF8}{gbsn}比如{\color{red}{里考塔芝士卷 (\textit{Ricotta cheese rolls})}}和{\color{red}{意式千层酥(\textit{Italian mille-feuille})}}一样。 \end{CJK} \\
    
    \bottomrule
    \end{tabular}}
    \caption{Prompting strategy examples (\textbf{Top}) and a source with cultural knowledge for En-Zh translation (\textbf{Bottom}).
    % \junjie{\binwei{Done}Confirm the prompts in self-correction and self-ranking}
    }
    \label{tab: Prompting Strategy}
    \vspace{-3mm}
\end{table*}

%% file: sections/05_results.tex
\section{Results and Analysis}
\label{sec:results}
% To investigate the cultural-awareness of NMT and LLM-based MT, we conduct automatic evaluations between LLM-based MT and popular NMT systems (\Scref{sec:auto_eval}) on six language pair translations. Diving into LLM-based MT, to explore the influence of prompting strategy on LLMs in culture-specific translation, we conduct automatic evaluations of LLM prompting strategies for English-Chinese translation (\Scref{sec:prompting_auto_eval}). Additionally, to  3) human evaluation on a subset of English-to-Chinese translation(\Scref{sec:human_evaluation}).
In this section, we 1) compare the CSI translation performances of LLM-based MT systems with that of NMT systems (\Scref{sec:auto_eval}); 2) evaluate dictionary-based terminology translation methods to explore if they work on CSI translations (\Scref{sec:traditional_translation}); 3) compare four prompting strategies of LLM-based MTs to explore how different prompts affect the LLMs' cultural awareness (\Scref{sec:prompting_auto_eval}); 4) conduct human evaluation to verify the correlation between automatic evaluation metrics and human assessment (\Scref{sec:human_evaluation}).

\subsection{Evaluating LLM-based MT v.s. NMT}
\label{sec:auto_eval}
To compare the cultural awareness of LLM-based MT systems with that of NMT systems, 
We employ two automatic metrics (CSI-Match and PTA) to evaluate 4 MT systems: the vanilla NLLB and Google Translate, and the BI prompting of LLaMA2 and GPT-3.5. 
\begin{figure}[h]
    \centering \includegraphics[width=0.45\textwidth]{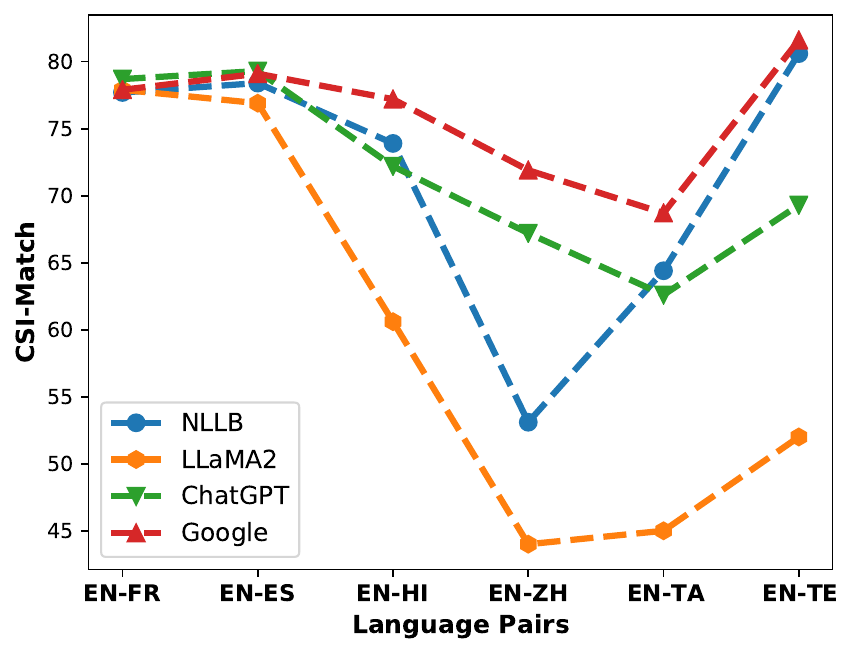}
    \caption{CSI-Match results on six language pairs.}
    \label{fig:cs-match-languages}
    \vspace{-5mm}
\end{figure}
% \paragraph{NMTs Have More Consistent Accuracy Across Languages than LLMs.}
\paragraph{NMTs Excel in CSI-Match on Low-resource Languages}
We evaluate the accuracy of CSI translations using CSI-Match across six language pairs, as shown in Figure \ref{fig:cs-match-languages}. For the two Romance languages (French and Spanish), the performances of four MT systems are quite similar. However, Google Translate generally exhibits more consistent performance in non-Romance languages compared to LLM-based MT. NLLB's poor performance on EN-ZH is due to its limited translation capacity on EN-ZH, as validated in previous benchmarking work \cite{aharoni-etal-2019-massively, akinade-etal-2023-varepsilon}. Specifically, NLLB and Google Translate outperform LLaMA2 and GPT-3.5 in translating three Indian languages: Hindi, Tamil, and Telugu. Additionally, for low-resource languages like Tamil and Telugu, the translation performance of LLM-MTs remains limited on traditional translation metrics (see Table \ref{tab:autoeval}). Therefore, LLM-MTs cannot yet be considered efficient multilingual translation tools for these low-resource languages.
% Additionally, we measure the CSI-Match of ChatGPT using basic instructions (BI), and Google Translate (the \textcolor[HTML]{d62728}{red} solid line). We find that Google Translate's performance across different languages is generally more consistent compared to ChatGPT. Specifically, Google Translate performs better in non-Latin languages such as Chinese, Hindi, Tamil, and Telugu, when compared to ChatGPT.
% \input{tables/13_understandability}

\begin{figure}[t]
    \centering \includegraphics[width=0.48\textwidth]{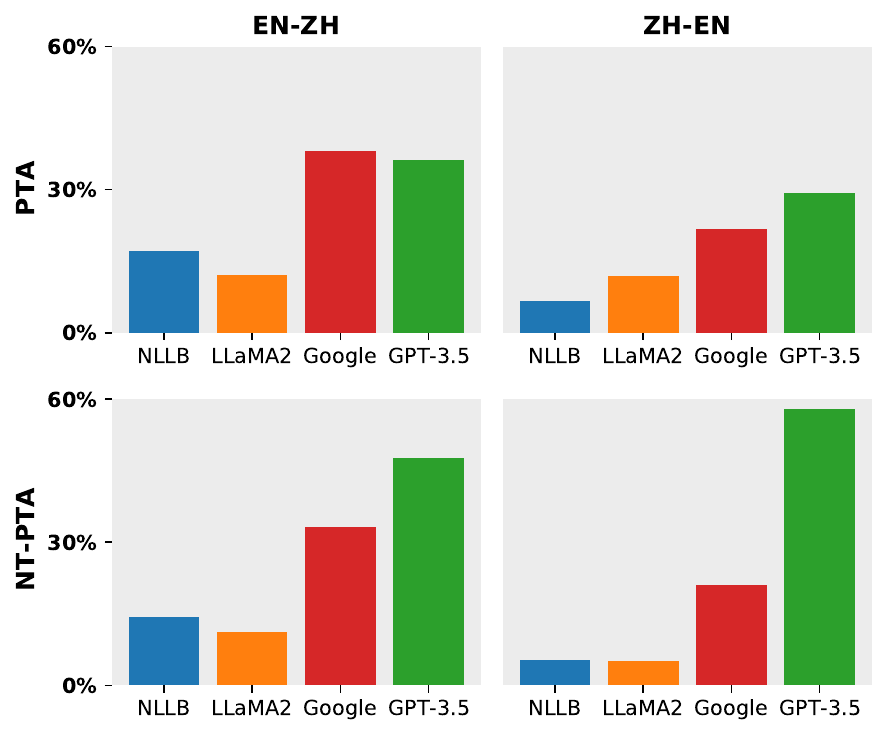}
    \caption{PTA results on English-Chinese translations.}
    \label{fig:pta-languages}
    \vspace{-3mm}
\end{figure}

% \paragraph{LLM Opens Opportunities to Pragmatic Translation Improvement.} 
\paragraph{GPT-3.5 Produces Better Pragmatic Translation on CSIs with No Established Translations.}
Given the cost of evaluation by commercial tools and human experts across all language pairs, we focus on English-Chinese in both translation directions when evaluating the pragmatic translation quality by PTA.
Figure \ref{fig:pta-languages} presents the PTA assessed by GPT-4o for four MT systems' output compared to the reference sentences. In addition to PTA across the entire dataset, we separately evaluate PTA of samples containing CSIs with no established translations (\textbf{NT-PTA}). Notably, GPT-3.5 performs better than any other MT systems on NT-PTA, which potentially suggests that the instruction-tuning of LLMs beyond the translation task may be beneficial for the model to generate free translations that are easily understood by the target culture, especially for non-translation CSIs.

\subsection{Evaluating Dictionary-based Methods}
\label{sec:traditional_translation}

\input{tables/22_dictionary_methods}
% To investigate whether traditional dictionary-based terminology translation methods can aid in translating culturally specific items (CSIs), we apply the Append and Replace methods (details in \Scref{sec: mt_systems}) to the open-source MT systems NLLB and LLaMA2 , and evaluate their translation quality. . 

%\paragraph{LLM-based MTs are Better at Leveraging External Knowledge.}
% \paragraph{Replacing CSI translations Works for NLLB, while Appending CSI Translation works for LLM-MT.}
% \paragraph{Traditional Dictionary-based Methods also Works for LLM-MT.}
\paragraph{LLaMA is More Robust at Leveraging CSI Dictionaries than NLLB.} We evaluate two dictionary-based terminology methods on NLLB and LLaMA for English-Chinese translations, as shown in Table \ref{tab:understandablity_dict_methods}. We find that straightforward strategies using dictionaries of CSIs, such as Replace and Append are effective for both NLLB and LLaMA on metrics that rely on string-matching (i.e. CSI-Match), as well as other semantic matching metrics (see Table \ref{tab:autoeval}). However, the appending strategy significantly benefits LLaMA more than NLLB. This suggests that LLaMA's ability for in-context learning and instruction-following enables the flexible integration of cultural knowledge at test time, a capability not present in traditional NMT systems like NLLB. Furthermore, traditional terminology translation methods can improve the PTA across the entire dataset. Without dictionaries, they still encounter challenges in improving the comprehensibility of translations that contain CSIs.

\subsection{Evaluating Prompting Strategies}
\label{sec:prompting_auto_eval}
\input{tables/22_prompting_PTA.tex}

\input{tables/17_prompting_result_examples}
LLM-based MTs open up the opportunity to incorporate free-form external knowledge to enhance the pragmatic translation quality of CSIs, especially for those without dictionaries. We explore various prompting strategies by integrating additional cultural knowledge, including dictionaries and explanations, to improve translations. We compare different prompting strategies for English-to-Chinese translations, employing 2-shot prompting approaches to obtain results from GPT-3.5 and LLaMA2.
% Especially for complex prompting strategies including CE, SE, and SR, we use 2-shot examples that are tailored for culturally-aware translation.
Table~\ref{tab:understandablity_prompting_methods} shows the evaluation results.

\paragraph{External CSI Knowledge Improves LLM-MT.}
% \paragraph{Prompting with CSI Explanations Improves LLM-based MT.}

When comparing the strategies of using external knowledge in prompts (i.e., CT and CE), we observe that LLMs can effectively leverage both direct translations and indirect explanations. Specifically, CT enhances the CSI-Match score for CSI translations. However, CT is not effective when CSIs have no existing translations. In contrast, the 2-shot CE approach using GPT-3.5 improves NT-PTA from 31.7 to 41.3 in English-to-Chinese translations. This suggests that CSI explanations can notably aid in translating CSIs, particularly those without well-known translations. For LLaMA, the PTA score is similar to the baseline (with differences of fewer than 10 examples out of 778 data points) due to the limitations of LLaMA2-7B's capacity for English-to-Chinese translations. However, CE and SE approaches with LLaMA still show an improvement in NT-PTA, indicating that LLaMA2-7B can also leverage external explanations to improve the translation quality of CSIs without existing direct translations.

\paragraph{LLMs' CSI Explanations Also Help.} We use SE to elicit LLMs' internal knowledge and find that the 2-shot SE approach with GPT-3.5 improves translation performance across all metrics for English-to-Chinese translations. This suggests that GPT-3.5 may already possess a significant amount of cultural knowledge about CSIs and can integrate this knowledge into the translation process. For LLaMA2, the PTA of SE is close to the baseline, and the improvement in NT-PTA is also limited, indicating that LLaMA2-7B may not have sufficient cultural knowledge of CSIs for English-to-Chinese translations.
% \begin{figure}[h]
%     \centering \includegraphics[width=0.35\textwidth]{figures/stats_prompting_stragies.pdf}
%     \caption{Translation Methods Statistics}
%     \label{fig:translation_strategies}
%     \vspace{-5mm}
% \end{figure}

\paragraph{Prompting with CSI Explanations Encourages LLMs to Do Free Translations.}In Table \ref{tab: Prompting Strategy Examples}, we provide examples of the CSI ``polenta'', an Italian corn porridge. Its Chinese translation on Wikidata is merely a transliteration. Under the CT strategy, GPT-3.5 directly copies this transliteration into the output, which may be considered correct but not comprehensible for native speakers of Chinese. In contrast, using the CE strategy, GPT-3.5 integrates the CSI explanation into the translation, freely translating the term ``corn porridge'' into Chinese. This makes it easier for readers to understand the nature of  ``polenta''. Furthermore, The SE strategy successfully generates an explanation for ``polenta'' and incorporates it into the translation as ``corn porridge'', which leads to better comprehension for Chinese native speakers, as is reflected in the ratings of GPT-4o and the human annotator.
% To investigate why explanation-based approaches can improve the translation accuracy of CSIs, we annotate the translation methods (literal translation or free translation) on a subset of 200 English-to-Chinese translation samples. The results, shown in Figure 
% \ref{fig:translation_strategies}, indicate that CE and SE tend to have more free translations than BI. Additionally, there are fewer entities identified as "Wrong" compared to BI. However, it is notable that CE and SE still adopt literal translation to a significant extent. We believe that exploring ways to increase the use of free translations is a promising direction for facilitating cross-culture communication in the future.

% On the other hand, SR proves beneficial on ChatGPT for CSI-Match but does not improve the understandability of CSI translations, indicating that LLMs inherently possess internal knowledge of low-frequency CSIs but still can not capture cultural nuances through the SR prompting strategy examined in this study. 
% In Table~\ref{tab: Prompting Strategy Examples}, 2-shot SR translates ``Polenta'' as ``cornmeal,'' similar to the baseline (BI). However, SE successfully translates it into ``corn porridge.'' This suggests that ChatGPT possesses background knowledge of Polenta but struggles to spontaneously elicit it to generate and select the most understandable translations.
% \junjie{SE is 2-shot, but SR here is zero-shot? The last sentence is not fair.}
\subsection{Human Evaluation}
\label{sec:human_evaluation}
\input{tables/19_metric_consistency_all}
To compare the consistency between automatic metrics and human evaluation, we conduct a human evaluation on a subset of 200 English-to-Chinese translations. We assess the outputs from eight MT systems: NLLB, LLaMA2, Google Translate, and GPT-3.5 across zero-shot BI, and two-shot BI, CT, CE, and SE settings. The native Chinese speaker evaluates the accuracy and PTA of the outputs. We then calculate Pearson's correlation coefficients between automatic evaluation metrics and human assessments. Specifically, we compare the performance of CSI-Match and PTA with four traditional automatic evaluation metrics: BLEU, BLEURT, COMET, and Exact-Match. The results, presented in Table \ref{tab: correlation}, indicate that CSI-Match exhibits the highest correlation with human accuracy, while GPT-4o PTA shows the highest correlation with human PTA. These findings suggest that CSI-Match and PTA are effective evaluation metrics for assessing the translation quality of CSIs, which can better capture cultural nuances than traditional metrics.

%% file: tables/22_dictionary_methods.tex
\begin{table}[h]
    \centering
    % \vspace{-3mm}
    \resizebox{0.49\textwidth}{!}{
    \begin{tabular}{lllcc}
    \toprule
    \textbf{Model} &  &\textbf{Method}&  \textbf{CSI-Match} & \textbf{PTA} \\
    \midrule
   \multirow{6}{*}{\textbf{NLLB}}& \multirow{3}{*}{EN-ZH} & \textbf{Vanilla}  & 53.1 & 17.1\\
    & &\textbf{Append} & 58.3$\greenuparrow$ & 17.7$\greenuparrow$ \\
    & &\textbf{Replace}& \textbf{78.7}$\greenuparrow$ & \textbf{20.5}$\greenuparrow$ \\
    \cmidrule{2-5}
    &\multirow{3}{*}{ZH-EN} &\textbf{Vanilla}  & 64.9 & 6.7  \\
    & &\textbf{Append} & 65.5$\greenuparrow$ & 4.6$\reddownarrow$ \\
    & &\textbf{Replace}& \textbf{79.8}$\greenuparrow$ & \textbf{9.2}$\greenuparrow$ \\
    \midrule
    % \textbf{LLaMA2} & 12.1 & 11.1 & 12.0 & 5.3\\
        % \textbf{LLaMA2-A} & 16.8 & 7.9 &17.5 & {\textcolor{red}{26.3}} \\
    % \textbf{LLaMA2-R} & 15.3 & 11.1 & 16.5 & 0 \\
    \multirow{6}{*}{\textbf{LLAMA2}}& \multirow{3}{*}{EN-ZH} & \textbf{Vanilla}  & 45.0 & 12.1 \\
    & &\textbf{Append} & \textbf{80.2}$\greenuparrow$ & \textbf{16.8}$\greenuparrow$ \\
    & &\textbf{Replace}& 67.1$\greenuparrow$ & 15.3$\greenuparrow$ \\
    \cmidrule{2-5}
    &\multirow{3}{*}{ZH-EN} &\textbf{Vanilla}  & 70.9 & 12.0 \\
    & &\textbf{Append} & \textbf{85.6}$\greenuparrow$ & \textbf{17.5}$\greenuparrow$ \\
    & &\textbf{Replace}& 80.6$\greenuparrow$ & 16.5$\greenuparrow$ \\
    \bottomrule
    \end{tabular}
    }
    \caption{Evaluation of traditional dictionary-based methods on English-Chinese translations. $\greenuparrow$/$\reddownarrow$ means the score is better or worse than the vanilla model.}
    \label{tab:understandablity_dict_methods}
     \vspace{-3mm}
\end{table}

%% file: tables/22_prompting_PTA.tex
\begin{table}[h]
    \centering
    % \vspace{-3mm}
    \resizebox{0.49\textwidth}{!}{
    \begin{tabular}{llccc}
    \toprule
    \textbf{Model} &\textbf{Method}&  \textbf{CSI-Match} & \textbf{PTA} & \textbf{NT-PTA} \\
    \midrule
   \multirow{4}{*}{\textbf{GPT-3.5}}& \textbf{BI}  & 66.2 & 33.2 & 31.7 \\
     &\textbf{CT} & \textbf{84.0}$\greenuparrow$ & 35.6$\greenuparrow$ & -\\
    &\textbf{CE}& 67.1$\greenuparrow$ & 35.8$\greenuparrow$ & \textbf{41.3}$\greenuparrow$ \\
    &\textbf{SE} & 67.7$\greenuparrow$ & \textbf{36.7}$\greenuparrow$ & 36.5$\greenuparrow$ \\
    % \cmidrule{2-6}
    % &\multirow{4}{*}{ZH-EN} &\textbf{BI}  & 76.0 & 28.4 & 57.9 \\
    % & &\textbf{CT} & \textbf{93.3}$\greenuparrow$ & 27.2$\reddownarrow$ & 42.1$\reddownarrow$ \\
    % & &\textbf{CE}& 78.1$\greenuparrow$ & 28.7$\greenuparrow$ & 36.8$\reddownarrow$ \\
    % & &\textbf{SE} & 82.3$\greenuparrow$ & \textbf{29.4}$\greenuparrow$ & \textbf{73.7}$\greenuparrow$ \\
    \midrule
    \multirow{4}{*}{\textbf{LLaMA2}} &\textbf{BI}  & 43.7 & 11.3 & 11.1 \\
     & \textbf{CT} & \textbf{82.2}$\greenuparrow$ & \textbf{16.6}$\greenuparrow$ & - \\
    & \textbf{CE}& 43.3$\reddownarrow$ & 10.8$\reddownarrow$ & \textbf{15.9}$\greenuparrow$ \\
    & \textbf{SE} & 47.8$\greenuparrow$ & 
    10.3$\reddownarrow$ & 12.7$\greenuparrow$ \\
    % \cmidrule{2-6}
    % & \multirow{4}{*}{EN-ZH} &\textbf{BI}  & 69.9 & 10.2 & 22.4 \\
    % & & \textbf{CT} & \textbf{91.5}$\greenuparrow$ & \textbf{17.2}$\greenuparrow$ & 15.8$\reddownarrow$ \\
    % & & \textbf{CE}& 83.9$\greenuparrow$& 14.8$\greenuparrow$ & \textbf{31.6}$\greenuparrow$ \\
    % & & \textbf{SE} & 71.5$\greenuparrow$ & 11.8$\greenuparrow$& 15.8$\reddownarrow$ \\
    \bottomrule
    \end{tabular}
    }
    \caption{Evaluation of different prompting strategies on English-to-Chinese translations. $\greenuparrow$/$\reddownarrow$ means the score is better or worse than the vanilla model.}
    \label{tab:understandablity_prompting_methods}
     \vspace{-3mm}
     % \binwei{Examples of LLaMA2, NT-U generations in appendix; Why 2-shot}
\end{table}

%% file: tables/17_prompting_result_examples.tex
\begin{table*}[]
    \centering
    \resizebox{\linewidth}{!}{
    \begin{tabular}{llcc}
    \toprule
    \textbf{Strategy}  & \textbf{Outputs} & \textbf{PTA of GPT-4o} & \textbf{PTA of Human}  \\
    \midrule
\textbf{BI} & \begin{CJK}{UTF8}{gbsn}就像意大利的{\color{blue}polenta concia}一样，它可以作为主菜食用。\end{CJK} & Lose & Lose\\
\midrule
\textbf{CT}& \begin{CJK}{UTF8}{gbsn}和在意大利一样，{\color{blue}波伦塔(\textit{transliteration})} 在意大利被当作主菜。\end{CJK} & Lose & Lose\\
\midrule
\textbf{CE} &\begin{CJK}{UTF8}{gbsn}就像意大利的{\color{blue}奶酪玉米粥(\textit{cheese corn porridge})}一样，它可以作为主菜食用。\end{CJK} & Win & Win\\
\midrule
\textbf{SE} &\begin{CJK}{UTF8}{gbsn}就像意大利的{\color{blue}奶酪玉米粥(\textit{cheese corn porridge})}一样，它可以作为主菜食用。\end{CJK} & Win & Win\\
& Explanation by GPT-3.5: Polenta is a traditional Italian dish that originated in Northern Italy. It is a  \\
& {\color{blue}type of porridge made from cornmeal}, and is similar in consistency to grits or cornmeal mush.\\
 \midrule \midrule   
    \textbf{Source} & Just like {\color{red}polenta} concia in Italy, it is eaten as a main dish.\\
    % {\color{red}{sfogliatelle}}.\\
    \textbf{Reference} & \begin{CJK}{UTF8}{gbsn}就像意大利的{\color{red}{玉米粥 (\textit{corn porridge})}}一样，它可以作为主菜食用。\end{CJK}\\
    \midrule
    \textbf{Knowledge} & Translation: \begin{CJK}{UTF8}{gbsn}{\color{blue}波伦塔 (\textit{transliteration})} \end{CJK}\\
    & \multicolumn{3}{l}{Explanation: Polenta is a dish of boiled cornmeal that was historically made from other grains.} \\
    &  The dish comes from Italy. It may be served served as \color{blue}{a hot porridge.}\\
    
    \bottomrule
    \end{tabular}}
    \caption{The output example of four prompting strategies on GPT-3.5 for En-Zh translation. 
    % \junjie{\binwei{Done}Confirm the prompts in self-correction and self-ranking}
    }
    \label{tab: Prompting Strategy Examples}
    \vspace{-5mm}
\end{table*}

%% file: tables/19_metric_consistency_all.tex
\begin{table}[h]
    \centering
    \resizebox{0.9\linewidth}{!}{
    \begin{tabular}{l|cc}
    \toprule
    \textbf{Metrics}  & \textbf{Human Acc.} & \textbf{Human PTA}\\
    \midrule
    \textbf{BLEU} & 79.6 & 86.2 \\
    \textbf{BLEURT} & 80.6 & 86.6 \\
    \textbf{COMET}&77.8 & 89.3 \\
    \textbf{Exact-Match} & 77.0 & 81.7 \\
    \textbf{CSI-Match}&\textbf{88.7} & 90.0\\
    \textbf{GPT-4o PTA} & 87.1 & \textbf{95.7} \\
    \bottomrule
    \end{tabular}
    }
    \caption{Pearson’s Coefficients between automatic metrics and human evaluation on CSI translation}
    \label{tab: correlation}
    \vspace{-3mm}
\end{table}

%% file: sections/08_conclusion.tex
\section{Conclusion}
\label{sec:conclusion}
% In this paper, we propose the first parallel corpus with rich CSI annotations for Cultural-Aware Machine Translation (CAMT), constructed automatically. We also devise two evaluation metrics, CSI-Match and PTA, which address cultural nuances beyond mere accuracy, particularly for terms lacking established translations. Using our dataset and metrics, we assess LLM-based MT systems and traditional NMTs, along with various cultural prompting strategies. The results indicate that LLM-based MT systems can effectively elicit cultural knowledge, enhancing the understandability of CSI translations. Despite the effectiveness, several challenging questions remain open. \binwei{TODO}How to more efficiently use external knowledge? How to better define understandability?

To advance culturally-aware machine translation, we curate a high-quality, diverse parallel corpus (CAMT) with rich CSI annotations in 6 language pairs using an automated pipeline. We introduce two evaluation metrics, CSI-Match and PTA, to assess translation quality concerning cultural nuances. Our evaluation of LLM-based MT and NMT systems using CAMT reveals that LLMs can effectively incorporate external cultural knowledge, enhancing the pragmatic translation quality of CSIs. Our work provides essential data sources and insights for advancing culturally-aware machine translation, laying the groundwork for future investigation in this field.
% First, it is non-trivial to incorporate cultural-specific information beyond a single entity such as discourse information. Besides, our prompting strategies leverage mostly external cultural knowledge in the form of texts. How to leverage multimodal knowledge from images and structured knowledge graphs to resolve cultural ambiguity deserves further investigation. 
% Finally, automatic evaluations of cultural nuances for LLM-based machine translation are also challenging, as LLMs tend to generate lengthy target-language explanations which may be different from translation references in parallel corpora.

% In this paper, we introduce a novel data curation pipeline aimed at constructing a culturally sensitive parallel corpus to assess the cultural awareness of MT systems. Additionally, we propose a reference-free metric to evaluate the understandability of translations for cultural-specific content by GPT-4. We also devise simple yet effective prompting strategies to enhance the understandability of LLM-based translations. Despite their effectiveness, several challenging questions remain open. First, integrating cultural-specific information beyond individual entities, such as discourse information, presents non-trivial challenges. Moreover, while our prompting strategies predominantly utilize external cultural knowledge in textual form, exploring the integration of multimodal knowledge from images and structured knowledge graphs to resolve cultural ambiguity warrants further investigation.

%% file: sections/09_limitations.tex
\section*{Limitations}
\label{sec:limitations}
\paragraph{Language Pairs in Evaluation} Our work takes a significant step toward toward understanding and evaluating the cultural awareness of machine translation on CSIs. We provide a culturally sensitive parallel corpus with rich annotations on cultural-specific items in six languages pairs. However, due to the cost of evaluation by commercial LLMs and human experts across all language pairs, we conduct parts of our experiments on English-Chinese translations, whose data quality is also verified by human experts. Building on our insights into English-Chinese translation, we hope to encourage future work to verify our findings on other language pairs, and we will release our code repository to streamline further investigations.

\paragraph{Cultural-Awareness Definition} 
In this study, we focus on the evaluation of cultural-specific items (CSIs). However, evaluating cultural awareness beyond individual entities also deserves further investigation. Besides CSIs, many other types of cultural errors persist in the translation process, such as those related to linguistic style and slang~\cite{hershcovich2022challenges}. Our work aims to mitigate cultural errors by starting with CSIs, promoting advancements in culturally-aware machine translation datasets, models, and evaluation methods. This is crucial for enabling machine translation to play a larger role in cross-cultural communications. 
\paragraph{Evaluation by LLM} 
Recent research has shown that GPT-4 demonstrates a high correlation with human experts in evaluating generation performance~\cite{rafailov2023direct, kocmi2023large,li2024translate}. However, using GPT-4 as an evaluator may still pose fairness issues due to internal biases and unbalanced language capabilities of LLMs. In this study, we aim to advance beyond traditional semantic alignment evaluation metrics to assess pragmatic translation quality in English-Chinese translations using GPT-4. Further investigation is needed to improve GPT-4's effectiveness as an translation evaluator.
\paragraph{Prompting strategies}
We only try 4 prompting strategies in our study, due to our work's focus on benchmarking the cultural awareness of current LLM-based MT systems. In the future, we'll test other methods, such as instruction tuning, to improve the performance of LLM-based MT.

\section*{Ethical Considerations}
\label{sec:ethical}
Although our study designs a suite of simple but effective prompting strategies to enhance the cultural awareness of LLM-based machine translation, we still observe the weakness of LLM-based translation on cultural concepts in certain regions (e.g., Asia) and hallucinations on low-frequency entities. Potential usage of these LLM translation outputs may still result in the spread of misinformation. Before deploying our methods to create reliable content such as creating translations of Wikipedia articles, practitioners should ensure another round of human post-editing. 
During the annotation process, the annotators (native speakers of the target languages) consist of the authors of this article, who know the goals of the study clearly.

%% file: sections/10_acknowledgement.tex
\section*{Acknowledgements}
We sincerely appreciate the valuable feedback provided by our reviewers, which greatly helped to improve the manuscript. BY and JH are supported by the Wisconsin Alumni Research Foundation. MJ is partially
supported by the National Science Foundation (IIS-2438420). The content is solely the responsibility of the authors and does not necessarily represent the official views of the National Science Foundation.

%% file: sections/00_appendix.tex
\section{Data Examples}
\label{sec:data_example}
In Table \ref{tab: data example}, we present a data example from the English-Chinese corpus. Each data point consists of a pair of sentences. We meticulously annotate all culture-specific items (CSI) within the sentences. For each culture-specific item, we provide information including its category, country of origin, translations in the target language, descriptions in both the source and target languages, and an explanation. 
To illustrate the challenges that cultural-specific items pose for current Machine Translation (MT) systems, we provide translations from both Google Translate and ChatGPT for this example. It is noted that both Google and ChatGPT erroneously rendered the Chinese translation of "Wiener Schnitzel" as "pork chops" instead of the correct translation, which is "steak". This misinterpretation not only misleads Chinese readers but also introduces confusion to the entire sentence, whose meaning is ``\textit{The Shanghai-style pork chops are a twist on Austria's national dish, Wiener fried pork chops, which are more street food than steak.}''
\input{tables/11_data_example}

\section{CSI vs. Wikiproject Mapping Table}
\label{sec:map}
\input{tables/07_mapping_table}
The mapping table between CSI definitions~(5 categories in total) and Wikiproject categories\footnote{\url{https://en.wikipedia.org/wiki/Wikipedia:WikiProject_Categories}}~(18 categories) are shown in Table \ref{tab: mapping table}. Additionally, we provide examples for each category to clarify the respective meanings. The tool we used for Wikiproject category classification is drafttopic\footnote{\url{https://github.com/wikimedia/drafttopic}}. 

\section{Wikipedia Parallel Corpus Collection}
\label{sec:parallel_corpus_collection}
To collect the English-Chinese parallel corpus from Wikipedia, we use the bilingual Wikipedia articles translated through Wikipedia's content translate tool\footnote{\url{https://en.wikipedia.org/wiki/Wikipedia:Content_translation_tool}}. This tool allows confirmed editors to translate Wikipedia articles from the source language to a target language with a machine translation system. By tracking their editing logs, we obtain the text triples consisting of the original text in a source language, the machine-translated text, and the human post-edited text in the target language. We then use a sentence alignment tool \texttt{bleu-align}\footnote{\url{https://github.com/rsennrich/Bleualign}}~\cite{sennrich-volk-2010-mt} to obtain a sentence-level parallel corpus. To obtain more language pairs, we reuse open-source data from OPUS, which includes Wikipediav1.0
\footnote{\url{https://opus.nlpl.eu/Wikipedia-v1.0.php}} 
for English-French and English-Spanish, as well as Samanantarv0.2
\footnote{\url{https://opus.nlpl.eu/Samanantar-v0.2.php}} 
for English-Hindi, English-Tamil, and English-Telugu.
% \section{Prompting Method Examples}
% \label{sec:prompting_examples}
% The examples of different prompting methods are shown in Table \ref{tab: Prompting Strategy}.
% \input{tables/01_examples}

\section{Data Characteristics}
\label{sec:stats}
\begin{figure}[h]
    \centering
\includegraphics[width=0.98\columnwidth]{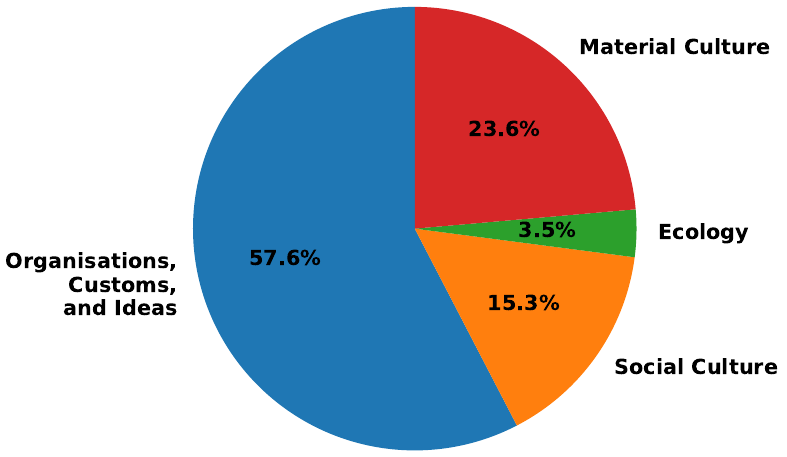}
    \caption{Category distribution on categories.}
    \label{fig:category_characteristics}
    % \vspace{-5mm}
\end{figure}

\begin{figure}[h]
    \centering
\includegraphics[width=\columnwidth]{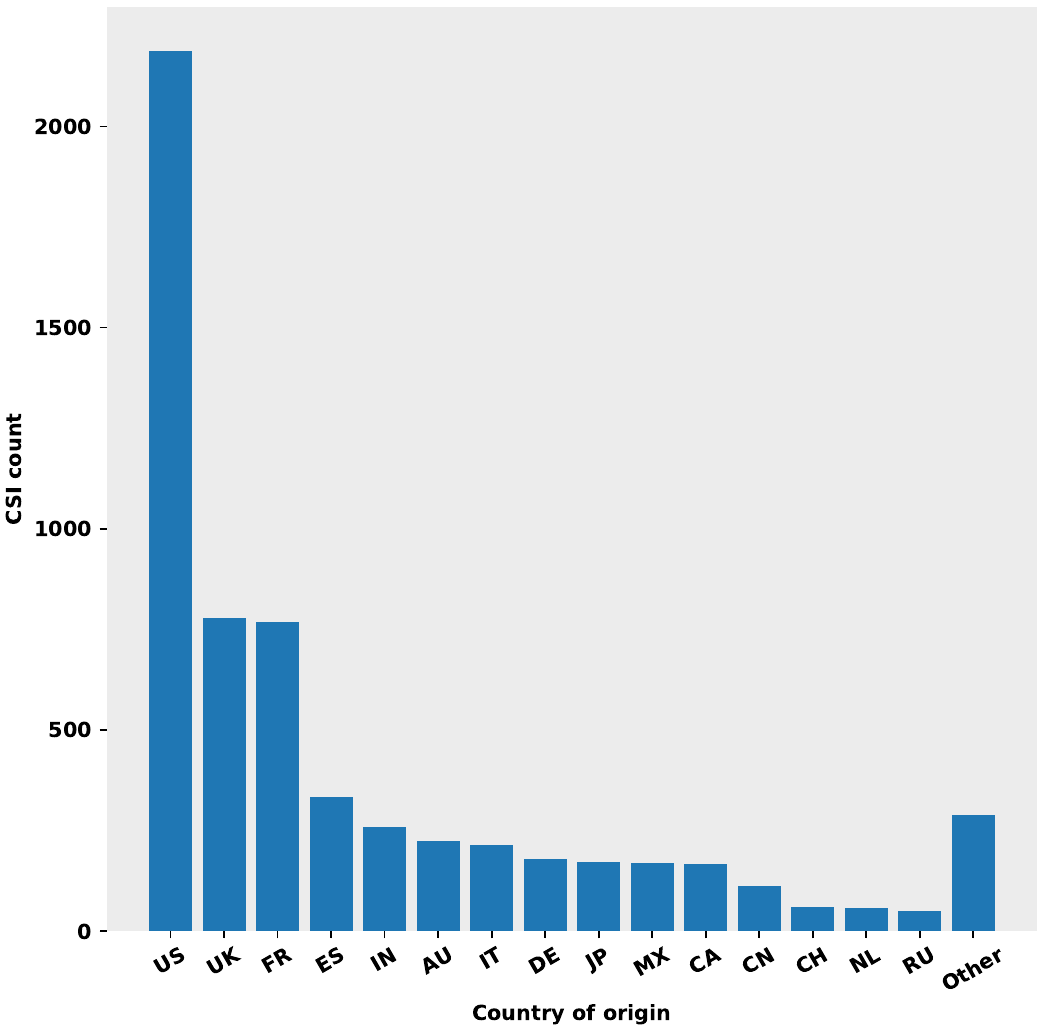}
    \caption{Data characteristics on regions.}
    \label{fig:region_characteristics}
    % \vspace{-5mm}
\end{figure}

% Culture is often associated with a specific region, and its expressions can vary significantly across different regions and categories. As a result, our dataset includes cultural-specific items from a diverse range of regions and categories to evaluate machine translation performance across a broad range of cultural contexts.
\paragraph{Data Statistics} Table~\ref{tab: dataset statistics} shows the statistics of our parallel corpora for the evaluation of MT systems on six language pairs. Particularly, for each language pair, we count the total number of detected CSIs by \textbf{CSIs Counts} and the number of unique CSIs by \textbf{CSIs Types}. It's noted that not all the CSIs have translations on Wikidata, so we determine the number of CSIs containing translations in WikiData by \textbf{CSI Translations}. Considering that many CSIs only exist within a specific culture group, which can't be located in the parallel corpus, CSIs that don't have a translation in other languages should take a higher proportion in the real-world corpus than in our dataset.
\paragraph{Data Diversity} Culture is intricately linked to specific regions, and its manifestations can exhibit substantial variations across diverse regions and categories. Therefore, our dataset encompasses culturally specific items sourced from a wide array of regions and categories. Figure \ref{fig:category_characteristics} shows the distribution of categories. Specifically, we mapped 18 Wikiproject categories into 5 culture categories. Since there is no Wikiproject category matching the CSI category \textit{gestures and habits}, we excluded this label from further consideration. Regarding the regions, we show the top 15 origin countries in our dataset in Figure \ref{fig:region_characteristics}. Among these regions, CSIs originating from English-speaking countries (e.g., the United States and the United Kingdom) have the highest representation. This is because we conduct entity-linking on the English source texts, resulting in a predominance of CSIs from English-speaking countries. However, the entity linking tool SLING\footnote{\url{https://github.com/google/sling}} is multilingual, making it feasible to use our pipeline to include more CSIs from non-English speaking countries. This inclusive approach allows us to comprehensively evaluate the performance of machine translation models across a broad spectrum of cultural contexts.

% %For the categories, we classify 18 Wikiproject categories into 5 culture categories as shown in Table \ref{tab: mapping table} (\ref{sec:map}). Since the category of \textit{gestures and habits} has no corresponding Wikiproject categories, we collect data from the remaining 4 categories. 
% \input{tables/00_data_stats}

% More data statistics of other language pairs is in the Appendix \Scref{sec:stats}. 

\section{Evaluation Prompts of GPT-4o}
\label{sec:evaluation_prompts}
It has been shown that GPT-4 can be an effective tool for evaluating the quality of generation tasks in DPO \cite{rafailov2023direct}. We apply a similar prompt method for the pragmatic translation quality evaluation of CSIs. The prompt is as follows, with the system output and reference randomly shuffled into choices A and B:
\begin{shaded}

{\noindent\textit{Assuming you’re a Chinese native speaker, which of the following translations has a more understandable translation in Chinese of following culture-specific item: ``Goubuli''? Please only compare the item\'s translation by ignoring the translation quality and length of the whole sentence.}

\noindent {\color{blue}<Source>}

\noindent\textit{Translation A:} {\color{blue}<A>}

\noindent\textit{Translation B:} {\color{blue}<B>}

\noindent\textit{FIRST, provide a one-sentence comparison of the two translation, explaining which you prefer and why. SECOND, on a new line, if the translations of cultural-specific items: ``Goubuli'' in "A" and "B" are different, state "A" or "B" to indicate your choice, otherwise, use "C" to indicate your choice. Your response should use the format:}

\noindent \textit{Comparison: <one-sentence comparison and explanation>}

\noindent \textit{Preferred: <"A" or "B" or "C">}}

\end{shaded}
For the human evaluation, we also use the same prompt as instructions to align the human evaluations with GPT-4o evaluations.
\section{Experiment Settings}
The experiment settings of different models included in our paper are as follows:
\begin{itemize}
    \item \textbf{NLLB} We use NLLB-200-1.3B-distilled\footnote{\url{https://github.com/facebookresearch/fairseq/tree/nllb?tab=readme-ov-file}} for our experiments. We use fairseq\footnote{\url{https://github.com/facebookresearch/fairseq}} to conduct the inference. The beam is set as 4, and the length penalty is set as 1.0.
    \item \textbf{LLaMA2} We use LLaMA-2-7B-hf\footnote{\url{https://huggingface.co/meta-llama/Llama-2-7b-hf}} for testing. The sampling is set as True, leading to a multinomial sampling searching method.
    \item \textbf{GPT-3.5} We examine version gpt-3.5-turbo-1106. We use the ChatCompletion\footnote{\url{https://platform.openai.com/docs/guides/text-generation/chat-completions-api}} API provided by OpenAPI For the generation, we set the parameters as default, for which the temperature is 1, top\_p is 1,  and frequency\_penalty as 0.
    \item \textbf{GPT-4o} For GPT-4o, we use the latest version gpt-4o-2024-05-13 on Microsoft Azure platform by ChatCompletion, and we set the parameters as following: the temperature is 0 for a stable generation, top\_p is 1,  and frequency\_penalty as 0.
    \item \textbf{Google translate} We call the Google Translate API\footnote{\url{https://cloud.google.com/translate/docs/reference/rest}} of Google Translate to get translations from it.
\end{itemize}

\section{Overall Automatic Evaluation}
\label{sec:overall_auto_metric}
\input{tables/02_overall_evaluation}

We evaluate the translation outputs using traditional automatic metrics such as BLEU~\cite{papineni-etal-2002-bleu}, BLEURT~\cite{sellam-etal-2020-bleurt}, and COMET~\cite{rei-etal-2020-comet}. To be consistent with the evaluation method of NLLB, we calculate spBLEU~\cite{goyal2022flores} for BLEU scores. In addition to traditional machine translation evaluation metrics, we also use CSI-Match to evaluate the translation quality of CSIs (described in \Scref{sec:culture_mt_evaluation}). Table~\ref {tab:autoeval} shows the results of eight MT systems across six language pairs in two directions.

As shown in Table~\ref {tab:autoeval}, both CSI dictionary incorporation (NLLB-A) and term replacement strategies (NLLB-R) enhance the translation quality of CSI for most language pairs, without significantly compromising the overall sentence translation regarding other metrics. Notably, NLLB-R outperforms other MT systems on CSI-Match, even including LLM-based MT. Interestingly, LLaMA2-7B shows an obvious drop in both traditional evaluation metrics and CSI-Match scores when translating English to three Indian languages and vice versa. One possible explanation is because of
the insufficient Indian data during the pre-training of LLaMA2. Both CSI-involving translation strategies are beneficial for LLaMA-based translation. In non-Romance languages (i.e., Chinese, Hindi, Tamil, and Telugu), LLaMA2-A tends to yield better performances, whereas LLaMA2-R performs better in Romance languages (i.e., French and Spanish), which potentially suggests that injecting cultural knowledge through code-switching similar Romance languages works better than distant languages for LLM-based models. Furthermore, we assess the translation performances of ChatGPT and Google Translate. Both MT systems exhibit commendable performance in CSI translation, with Google Translate demonstrating superior translation results. Notably, Google Translate showcases consistent translation abilities, particularly in handling relatively low-resource languages like Tamil and Telugu. 
\section{PTA Evaluation Results Across Languages}
\begin{figure}[h]
    \centering \includegraphics[width=0.48\textwidth]{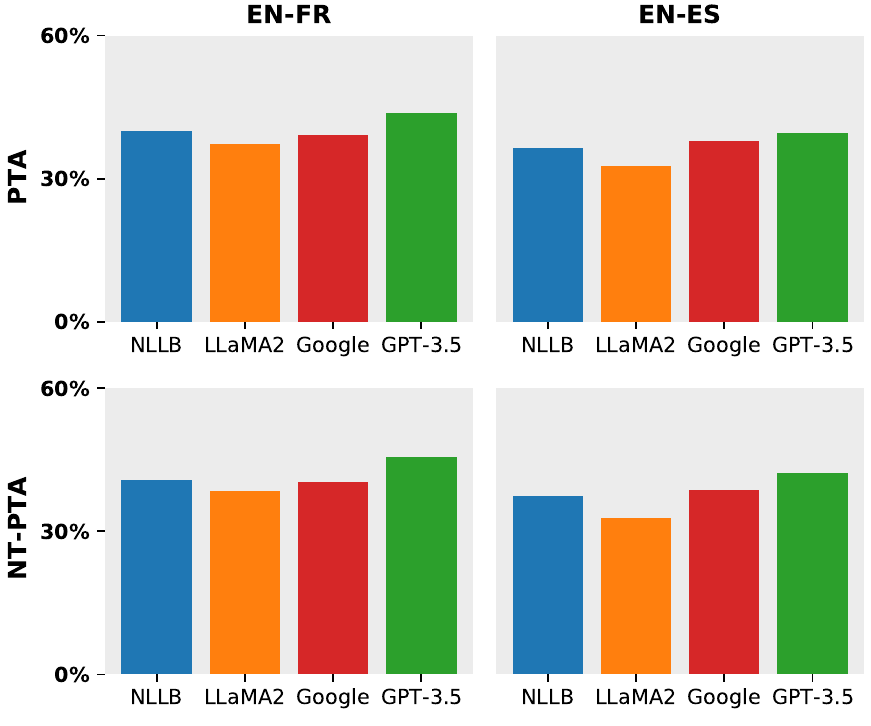}
    \caption{PTA results on En-Fr and En-Es translations.}
    \label{fig:pta-languages-additional}
    \vspace{-3mm}
\end{figure}

We evaluate the PTA of two more language pairs. The evaluation result is shown in Figure \ref{fig:pta-languages-additional}. As with CSI-match on these two languages, the PTA performances of the 4 MT systems are pretty close. However, GPT-3.5 still shows superior performance on PTA compared to NMTs, indicating that GPT-3.5 has better capabilities to generate free translations for CSIs which can be easily understood by native speakers in the target culture.
\section{Generation Data Examples of Non-translation CSIs}
Table~\ref{tab:NT_examples} shows the results of the four prompting strategies on the CSI ``milk toast'' from English to Chinese, which has no known translations. Under the BI strategy, GPT-3.5 translates the American breakfast dish as ``toast'', failing to capture the defining feature of the dish, which is that it is soaked in milk. CT similarly fails, yielding the term ``milk bread''. ``Milk'' as an adjective does not adequately describe the dish, and "bread" no longer specifies the toasted aspect. The former issue likewise arises with CE, a literal translation of ``milk toast''. In contrast, using SE, GPT-3.5 integrates the CSI explanation into the translation, freely translating the term as ``toasted bread soaked in milk''. This makes it easier for Chinese readers to understand the meaning of "milk toast", as is reflected in the ratings of GPT-4 and the human annotator.
\input{tables/21_NT_examples}

\section{Generation Examples of LLaMA}
\input{tables/22_llama_examples}
Table~\ref{tab:llama_examples} shows the results of four prompting strategies on the CSI ``burrito'' from English to Chinese, defined as a \textit{"flour tortilla wrapped into a sealed cylindrical shape around various ingredients."} Under the BI and CE strategies, LLaMA translates it as ``bag'' and ``shell'' respectively, failing to capture the essential feature of the dish, which is its rolled shape. The CT strategy successfully copies the dictionary translation. Interestingly, CE freely translates the word into "American southwest breakfast roll," accurately describing the food's shape. Additionally, CE prompts LLaMA to leverage related cultural knowledge to include the region description in the translation of the CSI.
% \section{Generation Data Examples of Explanation-based Methods}
% \label{sec:expr_examples}
% \section{Generation Data Examples of LLaMA2}

\section{Performance Across Regions}
\begin{figure}[h]
    \centering
\includegraphics[width=0.98\columnwidth]{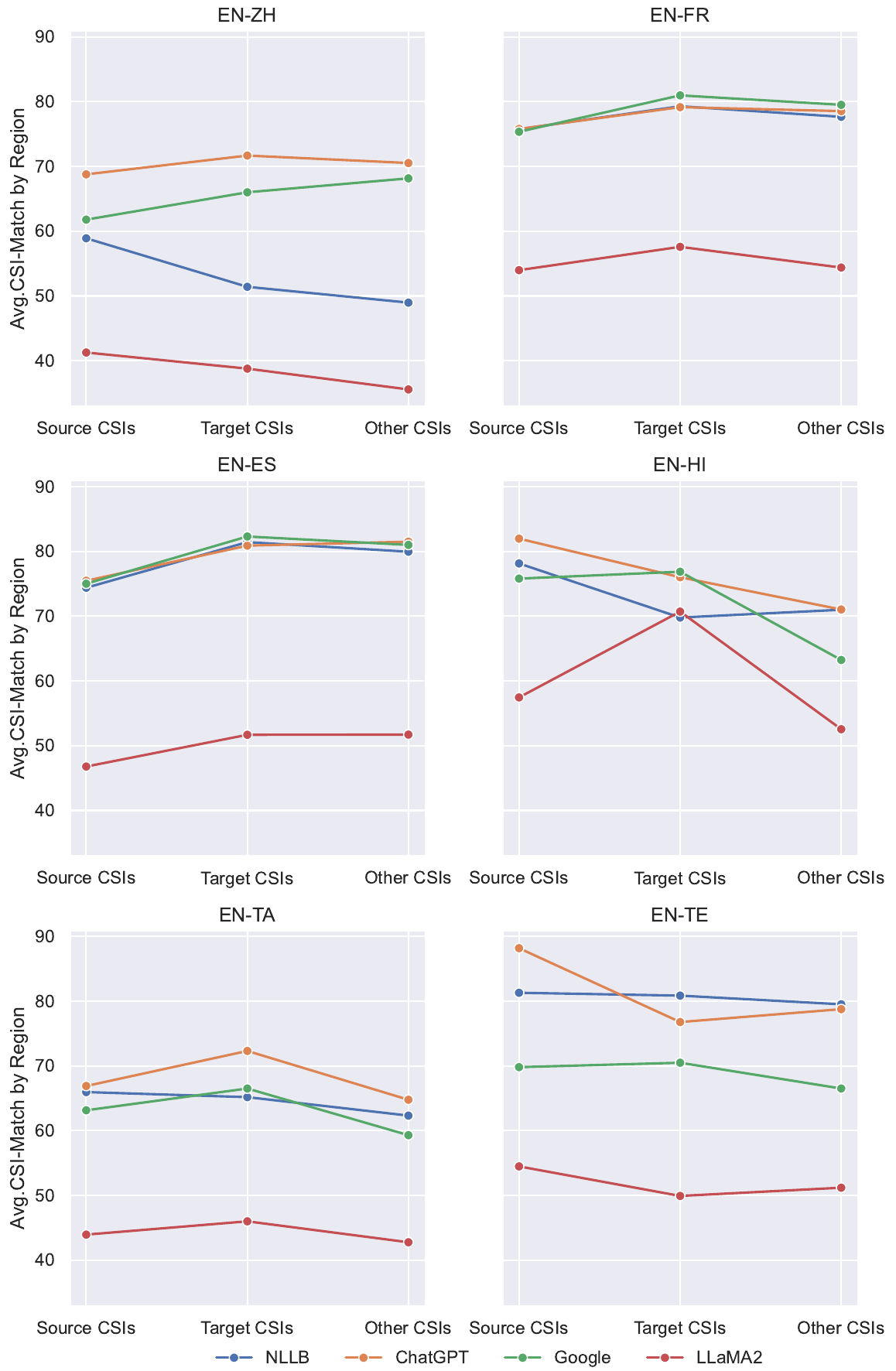}
    \caption{Avg.CSI-Match by regions.}
    % \vspace{-6mm}
    \label{fig:region_analysis}
\end{figure}
Culture is often associated with a specific region, and its expressions can vary significantly across different regions and categories. To gain a deeper understanding of the influence of region on CSI translation, we categorized CSI into three groups: CSI originating from countries primarily using the source language, countries predominantly using the target language, and countries utilizing languages other than the source and target languages. In the six groups of English-to-XX translations, we calculated the average CSI-Match values of these three CSI groups respectively, shown in Figure~\ref{fig:region_analysis}. 

Given that target CSIs must have the translation in the target language, translating target CSIs is akin to back translation. However, when translating the source CSI or other CSIs, the translation may either not exist in the target language or exist with lower word frequency. Consequently, the model is expected to yield better results for the target CSI.
Surprisingly, our analysis reveals that most models excel at translating the target CSI back into the target language in Romance languages (i.e., French and Spanish). Notably, Google Translate consistently achieves superior translations across all languages. ChatGPT demonstrates better translation performance in Chinese and Tamil, while LLaMA2 succeeds in Hindi and Tamil for target CSI translation. In contrast, traditional translation models NLLB struggle with all non-Romance languages, failing to outperform the source CSI translation. This suggests that LLMs may possess enhanced learning capabilities for translating culture-related content. However, it is important to note that the current translation performance is not consistently stable.

\section{Comparison of Translation Strategies}
\begin{figure}[t]
    \centering
\includegraphics[width=0.45\textwidth]{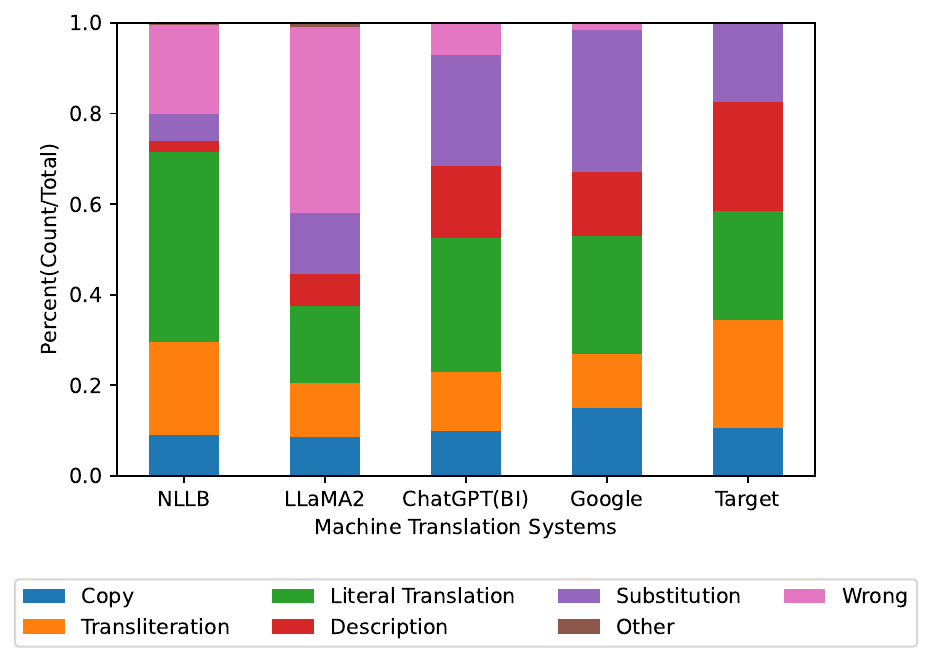}
    \caption{Percentage of translation strategies.}
    \label{fig:human_evaluation}
    % \vspace{-5mm}
\end{figure}
\paragraph{Translation Strategies} To explore potential factors benefiting pragmatic translation quality, we let a human annotator examine the models' translation strategies.  We categorize the translation strategies of CSIs based on prior translation theories~\cite{newmark1988textbook, persson2015culture}. These theories define different categorizations of strategies to improve the comprehensibility of CSIs while maintaining cultural integrity. We select 4 strategies that are common to our dataset. They're 
1) \textbf{Transliteration} that phonetically translates source CSIs; 2) \textbf{Literal translation} that directly translates word-by-word; 3) \textbf{Description} that integrates CSI descriptions of the CSIs into the translation; 4) \textbf{Substitution} that replace source CSIs by a semantically equivalent item in the target language; 5) \textbf{Copy} that directly copies the source language of CSI into the target language; 6) \textbf{Wrong} that indicates entirely incorrect translations; and 7) \textbf{Other} that employs other strategies in translation. Figure \ref{fig:human_evaluation} shows the ratio of each strategy in four MT systems. We find that models with higher PTA (e.g., ChatGPT and Google translate) use description and substitution at a significantly higher rate, indicating that these two strategies help improve the understanding of CSI for target-language speakers. 
Notably, LLaMA2 incorporates a higher frequency of substitution and description methods compared to traditional NLLB. However, this increased diversity in translation output comes at the cost of reduced stability in the outputs. As a result, LLaMA2 tends to yield more inaccurate translations, whereas NLLB relies more on Literal Translation and Transliteration to translate CSIs.
\begin{figure}[h]
    \centering
    \includegraphics[width=0.9\columnwidth]{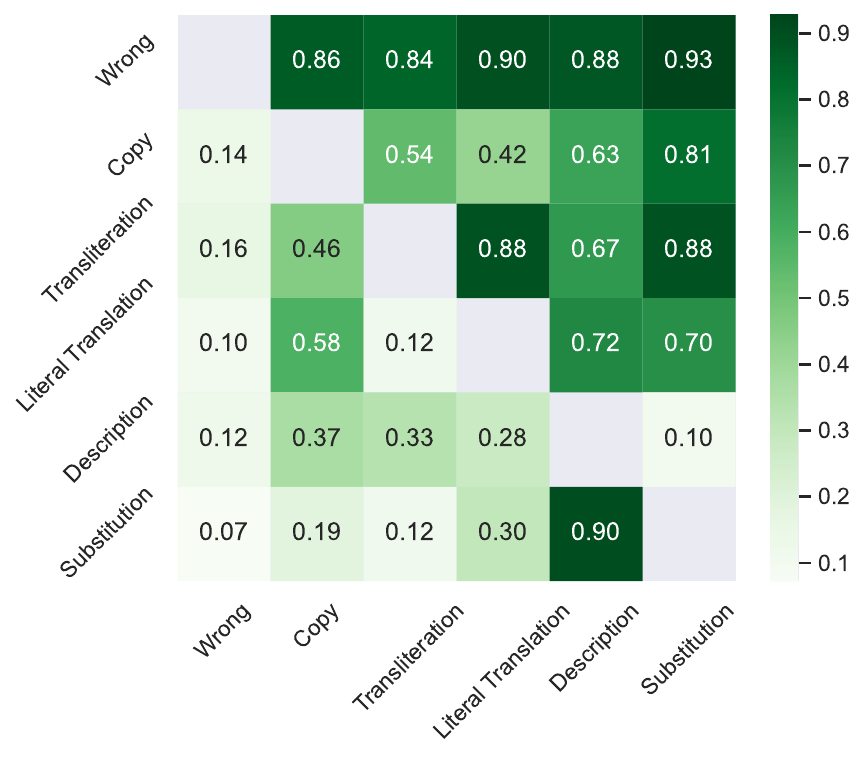}
    \caption{Comparison of Translation Strategies: The value in each grid represents the win rate of the method on the x-axis in comparison to the method on the y-axis.}
    \label{fig:strategy_comparison}
\end{figure}

In order to further compare the impact of different translation strategies on pragmatic translation quality, we analyzed the comparison between different translation strategies based on the ranking results of human evaluation. Specifically, we also rank the different translation strategies used by the MT systems according to the rank of the MT system's comprehensibility given by humans,  which is shown in Figure \ref{fig:strategy_comparison}. It's shown that the win rate of descriptions for all methods surpasses 0.5, and the win rate of substitutions, excluding descriptions, significantly exceeds 0.5. This implies that translations employing these two strategies are generally deemed more comprehensible by human annotators. Moreover, Literal Translation outperforms Transliteration, highlighting that transliteration may diminish the clarity of CSI in translation compared to a literal approach. Notably, the win rate of copying for both Literal Translation and Transliteration hovers around 50\%, indicating that these two methods may introduce confusion, and their readability underperforms directly copying the original word.

%% file: tables/11_data_example.tex
\begin{table}[h]
    \centering
    \resizebox{\linewidth}{!}{
    \begin{tabular}{ll}
    \toprule
    \textbf{Aspect} & \textbf{Content} \\
    \midrule
     \textbf{Source (EN)}    & The Shanghai-style Fried Pork Chop is\\
     &a modification from {\color{blue}{Wiener Schnitzel}}\\
     &the national dish of Austria, and a fried pork\\
     &chop is more a street food than a beef steak.\\
     \textbf{Target (ZH)}    & \begin{CJK}{UTF8}{gbsn}上海炸猪排的做法改良自奥地利国菜\end{CJK}\\
     & \begin{CJK}{UTF8}{gbsn}{\color{blue}{维也纳炸牛排 \textit{(Wiener fried steak})}}，而炸猪排\end{CJK} \\
     & \begin{CJK}{UTF8}{gbsn}与牛排不同，它显得更加市井。\end{CJK} \\
     \midrule
     \textbf{Cultural-Specific Item}  & Wiener Schnitzel\\
     \midrule
     \textbf{Category} & Culture.Food and drink \\
     \midrule
     \textbf{Country of Origin} & Austria \\
     \midrule
     \textbf{Translation (ZH)} & \begin{CJK}{UTF8}{gbsn}维也纳炸牛排 \textit{(Wiener fried steak)}\end{CJK} \\
     \midrule
     \textbf{Description (EN) } &  breaded veal schnitzel \\
     \textbf{Description (ZH) } &  \begin{CJK}{UTF8}{gbsn}面包屑小牛肉炸肉排 \end{CJK}\\
     \midrule
     \textbf{Explanation} & The entity, sometimes spelled Wienerschnitzel, \\
     &is a type of schnitzel made of a thin,\\
     &breaded, pan-fried veal cutlet. It is one of\\
     &the best known specialities of Viennese cuisine, \\
     &and one of the national dishes of Austria.\\
     \midrule
     \midrule
     \textbf{NLLB} & \begin{CJK}{UTF8}{gbsn}上海风格的炸猪肉切片是从奥地利\end{CJK}\\
     &\begin{CJK}{UTF8}{gbsn}国家菜的{\color{blue}{维也纳施尼切尔(\textit{transliteration})}}改制而成,\end{CJK}\\
     \midrule
     \textbf{LLaMA2} & \begin{CJK}{UTF8}{gbsn}上海炒猪排是一种来自奥地利\end{CJK}\\
     &\begin{CJK}{UTF8}{gbsn}的{\color{blue}{牛肉炒肉块 (\textit{Beef stir-fried cubes})}}的改良型，\end{CJK}\\
     &\begin{CJK}{UTF8}{gbsn}而炒猪排更像是一道街头小吃而非牛肉炒肉块\end{CJK}.\\
     \midrule
     \textbf{Google Translate} & \begin{CJK}{UTF8}{gbsn}海派炸猪排是奥地利国菜\end{CJK} \\
     & \begin{CJK}{UTF8}{gbsn} {\color{blue}{维也纳炸猪排 (\textit{Wiener fried pork chops})}}\end{CJK} \\
     & \begin{CJK}{UTF8}{gbsn}的改良版，炸猪排与其说是牛排，\end{CJK} \\
     & \begin{CJK}{UTF8}{gbsn}不如说是街头小吃。\end{CJK} \\
     \midrule
    \textbf{ChatGPT} & \begin{CJK}{UTF8}{gbsn}上海式炸猪排是从奥地利的国菜\end{CJK} \\
     & \begin{CJK}{UTF8}{gbsn} {\color{blue}{维也纳炸猪排 (\textit{Wiener fried pork chops})}}\end{CJK} \\
     & \begin{CJK}{UTF8}{gbsn}改编而来，而炸猪排比牛排，\end{CJK} \\
     & \begin{CJK}{UTF8}{gbsn}更像是街边食物。\end{CJK} \\
    \bottomrule
    \end{tabular}}
    \caption{A Data Example in the English-Chinese Corpus: in parentheses, we explain what the Chinese translation means.}
    \label{tab: data example}
    \vspace{-5mm}
\end{table}

%% file: tables/07_mapping_table.tex
\begin{table*}[h]
    \centering
    \resizebox{\linewidth}{!}{
    \begin{tabular}{lll}
    \toprule
     \textbf{CSI Category}    & \textbf{Wikiproject Category} & \textbf{CSI Example} \\
     \midrule
       Material Culture  &  Culture.Food and drink & \textit{Cotoletta (Italy)}\\
  &  Culture.Visual arts.Architecture  & \textit{the Summer Palace (China)}\\
    &  History and Society.Transportation & \textit{Kan-Etsu Expressway (Japan)}\\
    \midrule
    Social Culture & Culture.Sports  & \textit{RKC Waalwijk (Netherlands)}\\
   &Culture.Media.Entertainment & \textit{Far Rockaway (USA)}\\
    \midrule
    Organisations, & History and Society.Politics and government & \textit{Europe Ecology – The Greens (France)}\\
    Customs and Ideas &Culture.Philosophy and Religion & \textit{Fuller Theological Seminary (USA) }\\
    &Culture.Literature  & \textit{Der Spiegel (German)}\\
    &Culture.Visual arts.Visual arts* & \textit{The Headless Horseman Pursuing Ichabod Crane (USA)}\\
    &Culture.Visual arts.Fashion & \textit{Bottega Veneta (Italy)}\\
    &Culture.Visual arts.Comics and Anime & \textit{Dragon Ball (Japan)}\\
    &Culture.Performing arts & \textit{Just Dance (USA)}\\
    &Culture.Media.Music & \textit{Trident Studios (UK)}\\
    &Culture.Media.Films & \textit{A Few Good Men (USA)}\\
    &Culture.Media.Books & \textit{Moby-Dick (USA)}\\
    &History and Society.History & \textit{Tusculum (Italy)}\\
    \midrule
    Ecology & STEM.Biology & \textit{Kapok (Netherlands)}\\
    & Geography.Regions.* & \textit{Qualicum Beach (Canada)}\\
    \midrule
    Gestures and Habits & - \\
    \bottomrule
    \end{tabular}}
    \caption{CSI vs. Wikiproject mapping table.}
    \label{tab: mapping table}
\end{table*}

%% file: tables/02_overall_evaluation.tex
\begin{table*}[h!]
    \centering
    % \vspace{-4mm}
    \resizebox{\textwidth}{!}{
    \begin{tabular}{l|l|rrrr|rrrr}
    \toprule
    \textbf{Language Pair}& \textbf{Method}  & \textbf{BLEU} & \textbf{BLEURT} & \textbf{COMET} & \textbf{CSI-Match}  & \textbf{BLEU} & \textbf{BLEURT} & \textbf{COMET} & \textbf{CSI-Match}\\
    \midrule
    \multirow{9}{*}{\textbf{English-Chinese}} &&  \multicolumn{4}{c|}{EN-ZH}&\multicolumn{4}{c}{ZH-EN}\\
    
    &\textbf{NLLB} & 23.2 & 0.558 & 77.0 & 53.1 & 27.0 & 0.594 & 76.5 & 64.9 \\
    &\textbf{NLLB-A} & 17.3 & 0.447 & 62.8 & 58.3 & 21.9 & 0.531 & 69.4 & 65.5\\
    &\textbf{NLLB-R} & 23.9	& 0.555 & 76.8 & 78.7 & 25.3 & 0.591 & 75.4 & 79.8 \\
    &\textbf{LLaMA2} & 17.1 & 0.529 & 75.8 & 45.0 & 26.1 & 0.595 & 77.9 & 70.9\\
    &\textbf{LLaMA2-A} & 18.3 &	0.518 & 74.4 &  \textbf{80.2} & 29.1 & 0.629 & 79.0 & \textbf{85.6} \\
    &\textbf{LLaMA2-R} & 19.2 & 0.547 & 76.8 & 78.1 & 27.7 & 0.618 & 78.9 & 80.6 \\
    &\textbf{ChatGPT}  & 29.3 &	0.642 & 82.9 & 67.2 & 32.5 & 0.668 & 80.9 &77.7\\
    &\textbf{Google} & 38.3	& 0.679 & 84.2 & 71.9 & 41.1 & 0.697 & 82.2 &79.5\\
    \midrule
    
    \multirow{9}{*}{\textbf{English-French}}&&  \multicolumn{4}{c|}{EN-FR}&\multicolumn{4}{c}{FR-EN}\\
    &\textbf{NLLB} & 37.4 &	0.585 &	78.3 & 77.7 &	36.3&0.634&	77.8&	88.9\\
    &\textbf{NLLB-A}& 37.4 & 0.582 & 78.0 & 77.4 &35.4&0.628&77.2	&88.3 \\
    &\textbf{NLLB-R}& 37.0	&0.577&	77.8&\textbf{92.6}&36.6&0.635&77.7& 92.1 \\
    &\textbf{LLaMA2}& 34.4 & 0.558 & 76.2 & 77.9 & 27.6 &0.462&	66.0 & 72.5	 \\
    &\textbf{LLaMA2-A}& 35.0 & 0.568 & 67.8 & 82.7 & 34.7 & 0.633 & 77.7 & 92.8 \\
    &\textbf{LLaMA2-R}& 34.7 & 0.550 & 75.6 & 89.8 & 30.2 & 0.620 &76.8&\textbf{93.3} \\
    &\textbf{ChatGPT}  & 36.2 &	0.594 &	78.9 & 78.7  &37.6	&0.629&	77.8 & 89.8	\\
    &\textbf{Google} & 31.1 & 0.573 & 77.5 & 77.9 &36.1&0.632&77.2&88.9	 \\
    \midrule
    
    \multirow{9}{*}{\textbf{English-Spanish}} &&  \multicolumn{4}{c|}{EN-ES}&\multicolumn{4}{c}{ES-EN}\\
    &\textbf{NLLB} & 48.8 & 0.707 & 83.4 & 78.4 & 50.7 & 	0.718& 	83.5& 	90.3\\
    &\textbf{NLLB-A} & 48.3	& 0.705	& 83.3 & 78.7 & 49.9& 0.713& 83.0	& 90.8\\
    &\textbf{NLLB-R} & 47.9& 0.696& 82.9& \textbf{94.0}& 50.9& 0.717& 83.4& \textbf{95.2} \\
    &\textbf{LLaMA2} & 44.6& 0.679 & 81.6&76.9  & 45.3 & 0.704 &82.6 &89.9	\\
    &\textbf{LLaMA2-A} & 44.5 & 0.674 & 81.2 & 82.0 & 46.6 & 0.706 & 82.7&93.3 \\
    &\textbf{LLaMA2-R} & 44.5 & 0.675 & 81.3 & 93.0 &44.2&0.702&82.3&94.7	 \\
    &\textbf{ChatGPT}  & 47.9&	0.711 &	83.6 & 79.3 &50.7&	0.712	&83.4 &92.3	\\
    &\textbf{Google} & 42.9	& 0.704	&82.0& 79.1 &49.8	&0.722	&83.0&	90.9 \\
    \midrule
    
    \multirow{9}{*}{\textbf{English-Hindi}} &&  \multicolumn{4}{c|}{EN-HI}&\multicolumn{4}{c}{HI-EN}\\
    &\textbf{NLLB}     & 32.8 & 0.637 & 0.747 & 73.9 & 38.4 & 0.683 & 83.5 & 93.7 \\
    &\textbf{NLLB-A}   & 32.1 & 0.630 & 0.734 & 78.7 & 38.4 & 0.676 & 82.8 & 90.9 \\
    &\textbf{NLLB-R}   & 32.9 & 0.639 & 0.748 & \textbf{83.6} & 38.4 & 0.684 & 83.5 &	\textbf{98.3} \\
    &\textbf{LLaMA2}   & 7.3 & 0.438 & 0.493 & 60.6 & 12.6 & 0.441 & 63.8 & 67.5 \\
    &\textbf{LLaMA2-A} & 7.0 & 0.439 & 0.493 & 81.9 & 18.9 & 0.546 & 74.3 & 79.1\\
    &\textbf{LLaMA2-R} & 8.2 & 0.444 & 0.502 & 80.0 & 14.8 & 0.480 & 67.7 &	67.7 \\
    &\textbf{ChatGPT}  & 24.1 & 0.592 &	0.701 &	72.2 & 32.0 & 0.649 & 81.4 & 93.4	\\
    &\textbf{Google}   & 33.8 & 0.651 & 0.753 & 77.2 & 39.9 & 0.690 & 83.0 & 94.5 \\
    \midrule
    
    \multirow{9}{*}{\textbf{English-Tamil}} &&  \multicolumn{4}{c|}{EN-TA}&\multicolumn{4}{c}{TA-EN}\\
    &\textbf{NLLB} & 25.8&0.706&83.0&64.4& 31.5&	0.645&80.4	&92.1 \\
    &\textbf{NLLB-A} & 25.5	&0.698&82.3&70.5&31.0&	0.639&79.8&	94.9 \\
    &\textbf{NLLB-R} & 25.8&0.707&82.9&81.6&	31.5&0.645&80.4&\textbf{97.9} \\
    &\textbf{LLaMA2} & 1.7&0.309&39.6& 44.0 &4.3&0.331&51.9& 54.4 \\
    &\textbf{LLaMA2-A} &1.2&0.341&39.6&\textbf{88.9} & 9.1 & 0.417 & 61.5 & 87.0\\
    &\textbf{LLaMA2-R} &1.4&0.321&40.0&73.9&5.1&0.305&53.7&68.7	 \\
    &\textbf{ChatGPT}  & 10.4&0.496&62.5& 62.6 &16.9&0.539&	72.2&87.4	\\
    &\textbf{Google} & 26.9	&0.712	&82.7&	68.7&31.4&0.649&80.4&91.9 \\
    \midrule
    
    \multirow{9}{*}{\textbf{English-Telugu}} &&  \multicolumn{4}{c|}{EN-TE}&\multicolumn{4}{c}{TE-EN}\\
    &\textbf{NLLB} & 31.4&0.628&81.1&80.6 &34.7&0.643&81.2&87.3 \\
    &\textbf{NLLB-A} & 29.2	&0.624&	80.4&	81.9&34.5&0.635	&80.5	&89.8 \\
    &\textbf{NLLB-R} & 31.4 & 0.628&81.1&	\textbf{89.8}	&	34.7&	0.643& 81.2&\textbf{94.7} \\
    &\textbf{LLaMA2} & 3.2 & 0.207 & 41.0 & 52.0 & 0.9 & 0.190 & 41.6 & 33.2 \\
    &\textbf{LLaMA2-A} &4.0&0.244&42.3&88.8 & 5.8 & 0.356 & 56.7 & 78.0\\
    &\textbf{LLaMA2-R} &4.0&0.237&42.0&77.0&2.3&0.269&48.8&44.1	 \\
    &\textbf{ChatGPT}  &16.8 & 0.484  & 67.3	& 69.3 &23.3&0.567&74.8& 78.8	 \\
    &\textbf{Google} & 32.7	&0.635&81.0&81.6&34.9&0.653&81.2&	89.5 \\
    \bottomrule
    \end{tabular}
    }
    \caption{Automatic evaluation of all MT methods on six language pairs from both translation directions.}
    \label{tab:autoeval}
    % \vspace{-5mm}
\end{table*}

%% file: tables/21_NT_examples.tex
\begin{table*}[h]
    \centering
    \resizebox{\linewidth}{!}{
    \begin{tabular}{llcc}
    \toprule
    \textbf{Strategy}  & \textbf{Outputs} & PTA of GPT-4o & PTA of Human \\
    % \midrule
% \textbf{Example 1:} \\
\midrule
\textbf{BI} & \begin{CJK}{UTF8}{gbsn}该书中最终包含了1850种食谱，其中有{\color{blue}烤面包(\textit{toast})}。\end{CJK} & Lose & Lose \\
\midrule
\textbf{CT}& \begin{CJK}{UTF8}{gbsn}该书最终收录了1850个食谱，比如{\color{blue}{牛奶面包(\textit milk bread)}}。\end{CJK} & Lose & Lose \\
\midrule
\textbf{CE} & \begin{CJK}{UTF8}{gbsn}这本书最终包含了1850个食谱，其中有{\color{blue}牛奶土司(\textit literal translation})。\end{CJK} & Same & Same \\
\midrule
\textbf{SE} &\begin{CJK}{UTF8}{gbsn}该书最终包含了1850个配方，其中有{\color{blue}烤面包浸牛奶(\textit toasted bread soaked in milk})。\end{CJK} & Win & Win\\

 \midrule \midrule   
    \textbf{Source} & The book eventually contained 1,850 recipes including {\color{red} milk toast}. \\
    % {\color{red}{sfogliatelle}}.\\
    \textbf{Reference} & \begin{CJK}{UTF8}{gbsn}书中收录了1850个食谱，其中有{\color{red}{牛奶土司 (\textit literal Translation})}。\end{CJK}\\
    \midrule
    \textbf{Knowledge} & Translation: No existing Chinese translations\\
    & \multicolumn{3}{l}{Explanation: the entity is a breakfast dish consisting of {\color{blue}toasted bread in warm milk}, typically with sugar and butter...}\\
    % & served as \color{blue}{a hot porridge.}\\
    % \midrule
    % \textbf{Example 2:} \\
% \midrule
% \textbf{BI} & \begin{CJK}{UTF8}{gbsn}该书中最终包含了1850种食谱，其中有{\color{blue}烤面包(\textit{toast})}。制作韩式塔克斯的想法是在找不到{\color{blue}炭烤肉塔克斯(\textit(grilled meat})}后，由他提出的。\end{CJK} & Lose & Lose \\
% \midrule
% \textbf{CT}& \begin{CJK}{UTF8}{gbsn}该书最终收录了1850个食谱，比如{\color{blue}{牛奶面包(\textit milk bread)}}。\end{CJK} & Lose & Lose \\
% \midrule
% \textbf{CE} & \begin{CJK}{UTF8}{gbsn}这本书最终包含了1850个食谱，其中有{\color{blue}牛奶土司(\textit literal translation})。\end{CJK} & Same & Same \\
% \midrule
% \textbf{SE} &\begin{CJK}{UTF8}{gbsn}该书最终包含了1850个配方，其中有{\color{blue}烤面包浸牛奶(\textit toasted bread soaked in milk})。\end{CJK} & Win & Win\\

%  \midrule \midrule   
%     \textbf{Source} & The idea of making Korean tacos came to him after an unsuccessful search of {\color{red}carne asada} tacos. \\
%     % {\color{red}{sfogliatelle}}.\\
%     \textbf{Reference} & \begin{CJK}{UTF8}{gbsn}他在寻找 {\color{red}{carne asada (\textit Copy})} 墨西哥夹饼失败后萌生了研发韩式墨西哥夹饼的想法。\end{CJK}\\
%     \midrule
%     \textbf{Knowledge} & Translation: No existing Chinese translations\\
%     & \multicolumn{3}{l}{Explanation: the entity is {\color{blue}grilled and sliced beef}, usually chuck steak (known as Diezmillo in Spanish)...}\\
%     % & served as \color{blue}{a hot porridge.}\\
    \bottomrule
    
    \end{tabular}}
    \caption{Non-translation CSI output examples of prompting strategy on GPT-3.5 and a source-reference sentence pair with cultural knowledge for En-Zh translations.
    }
    \label{tab:NT_examples}
    % \vspace{-3mm}
\end{table*}

%% file: tables/22_llama_examples.tex
\begin{table*}[h]
    \centering
    \resizebox{\linewidth}{!}{
    \begin{tabular}{llcc}
    \toprule
    \textbf{Strategy}  & \textbf{Outputs} & PTA of GPT-4o & PTA of Human \\
    % \midrule
% \textbf{Example 1:} \\
\midrule
\textbf{BI} & \begin{CJK}{UTF8}{gbsn}爱尔兰早餐卷的制作方式类似于一顿{\color{blue}早餐休闲袋(\textit{breakfast relaxing bag})}。\end{CJK} & Lose & Lose \\
\midrule
\textbf{CT}& \begin{CJK}{UTF8}{gbsn}爱尔兰早饭卷饼是与{\color{blue}{墨西哥卷饼(\textit copy the dictionary)}}准确相似的。\end{CJK} & Win & Win \\
\midrule
\textbf{CE} & \begin{CJK}{UTF8}{gbsn}爱尔兰早餐卷的制作方式与{\color{blue}美国西南部的早餐卷(\textit American southwest breakfast roll)}一样。\end{CJK} & Win & Win \\
\midrule
\textbf{SE} &\begin{CJK}{UTF8}{gbsn}爱尔兰的早餐卷是和{\color{blue}早餐壳(\textit breakfast shell})。\end{CJK} & Win & Win\\

 \midrule \midrule   
    \textbf{Source} & The breakfast roll of Ireland is prepared similarly to a breakfast {\color{red}burrito}. \\
    % {\color{red}{sfogliatelle}}.\\
    \textbf{Reference} & \begin{CJK}{UTF8}{gbsn}爱尔兰的早餐面包卷制作方法亦类似于此 {\color{red}\textit{(Replace the term with it)}}。\end{CJK}\\
    \midrule
    \textbf{Knowledge} & Translation:\begin{CJK}{UTF8}{gbsn}墨西哥卷饼\end{CJK} \\
    & \multicolumn{3}{l}{Explanation: The entity is a dish in Mexican and Tex-Mex cuisine, consisting of } \\
    & {\color{blue} a flour tortilla wrapped into a sealed cylindrical shape around various ingredients...}\\
    \bottomrule
    \end{tabular}}
    \caption{CSI output examples of prompting strategy on LLaMA and a source-reference sentence pair with cultural knowledge for En-Zh translations.
    }
    \label{tab:llama_examples}
    % \vspace{-3mm}
\end{table*}